\documentclass{article}

\usepackage[final]{neurips_2024}

\usepackage{natbib}
\setcitestyle{numbers,square}
\usepackage{wrapfig}
\usepackage{amsmath}        
\usepackage[utf8]{inputenc} 
\usepackage[T1]{fontenc}    
\usepackage{hyperref}       
\usepackage{url}                 
\usepackage{amsfonts}      
\usepackage{amssymb}
\usepackage{nicefrac}       
\usepackage{microtype}      
\usepackage{xcolor}         
\usepackage{graphicx}
\usepackage{multirow}
\usepackage{booktabs}
\usepackage{tabularx}
\usepackage{subcaption}  
\usepackage{makecell}
\usepackage{enumitem}
\usepackage{adjustbox}
\usepackage{tcolorbox}
\usepackage{dialogue}
\usepackage{longtable}
\usepackage{colortbl}
\usepackage{longtable}

\title{ElectionSim: Massive Population Election Simulation Powered by Large Language Model Driven Agents}

\author{Xinnong Zhang\textsuperscript{\rm 1}\footnotemark[2],
        Jiayu Lin\textsuperscript{\rm 1}\footnotemark[2],
        Libo Sun\textsuperscript{\rm 1}\footnotemark[2],
        Weihong Qi\textsuperscript{\rm 2},
        Yihang Yang\textsuperscript{\rm 1},
        Yue Chen\textsuperscript{\rm 1},
        \\
        \textbf{Hanjia Lyu\textsuperscript{\rm 3},
        Xinyi Mou\textsuperscript{\rm 1},
        Siming Chen\textsuperscript{\rm 1}\footnotemark[3],
        Jiebo Luo\textsuperscript{\rm 3}\footnotemark[3],
        Xuanjing Huang\textsuperscript{\rm 1},
        Shiping Tang\textsuperscript{\rm 1},}
        \\
        \textbf{Zhongyu Wei\textsuperscript{\rm 1}\footnotemark[3]}
        \\
        \normalsize\textsuperscript{\rm 1}{Fudan University}, 
        \normalsize\textsuperscript{\rm 2}{Indiana University}, 
        \normalsize\textsuperscript{\rm 3}{University of Rochester}\\
        \normalsize\texttt{\{xnzhang23, jiayulin24, lbsun23, yhyang24, chen\_y24\}@m.fudan.edu.cn},\\
        \normalsize\texttt{\{xymou20, simingchen, xjhuang, twukong, zywei\}@fudan.edu.cn},\\
        \normalsize\texttt{wq3@iu.edu, hlyu5@ur.rochester.edu, jluo@cs.rochester.edu}\\ 
        \hyperref[http://www.fudan-disc.com/sharedtask/electionsim/]{Project Page: \texttt{\textcolor{blue}{http://www.fudan-disc.com/electionsim}}}
}

\begin{document}

\maketitle
\begin{abstract}
The massive population election simulation aims to model the preferences of specific groups in particular election scenarios. It has garnered significant attention for its potential to forecast real-world social trends. Traditional agent-based modeling (ABM) methods are constrained by their ability to incorporate complex individual background information and provide interactive prediction results. In this paper, we introduce \textbf{ElectionSim}, an innovative election simulation framework based on large language models, designed to support accurate voter simulations and customized distributions, together with an interactive platform to dialogue with simulated voters. We present a \textbf{million-level voter pool} sampled from social media platforms to support accurate individual simulation. We also introduce \textbf{PPE}, a poll-based presidential election benchmark to assess the performance of our framework under the U.S. presidential election scenario. Through extensive experiments and analyses, we demonstrate the effectiveness and robustness of our framework in U.S. presidential election simulations.
\end{abstract}

\footnotetext[2]{These authors contribute equally to this work.}
\footnotetext[3]{Corresponding authors.}

\begin{figure}[h]
    \centering
    \includegraphics[width=0.95\linewidth]{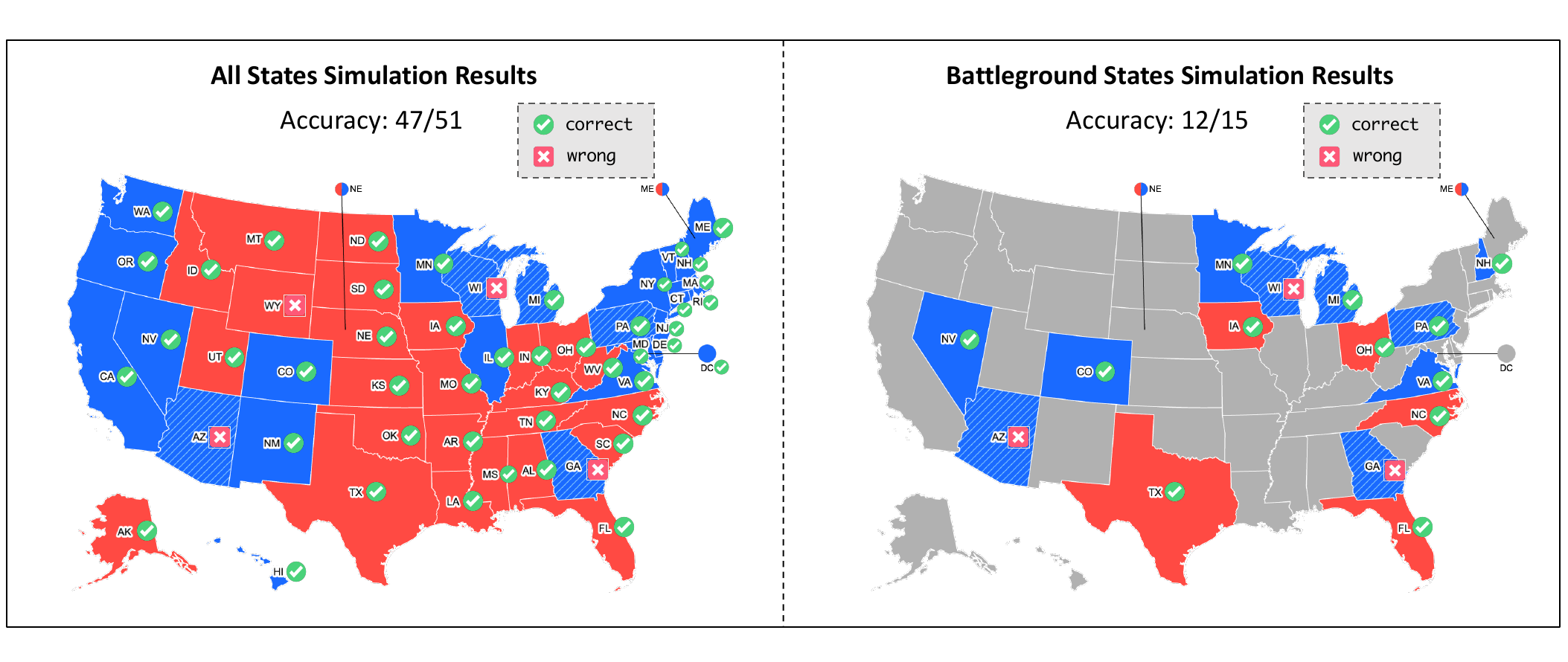}
    \caption{Simulation results of the 2020 Presidential Election. The colors represent the real-world results and the marks represent the simulation results accuracy.}
    \label{fig:head pic}
\end{figure}

\section{Introduction}\label{sec:intro}
Massive Population Election Simulation aims to simulate election events at a large population scale, which has been of vital importance in forecasting potential real-world political trends and capturing specific groups' preferences on particular topics or special events~\cite{hoey2018artificial,muric2022large,chopraagenttorch}. Previous works have demonstrated that using mathematical/statistical models can transform traditional political and sociological analysis paradigms~\cite{gao2022forecasting, mou2024unveiling}.

Currently, the mainstream method for election simulation is agent-based modeling (ABM)~\cite{schelling1969models,macal2009agent,jusup2022social,chuang2023computational}, which employs heuristic-like rules or mathematics functions to simulate the actions of individuals~\cite{tang2024idea}., and then scales up these actions to forecast the collective result.
The ABM approach, despite its merits, presents two major limitations. First, the integration of individuals' complex background information into the framework of ABMs is a non-trivial task. Second, the establishment of interaction human-agent interfaces to elucidate predictive results remains a significant challenge~\cite{choi2021theory, crooks2008key}. These two shortcomings can result in predictions that are less accurate and less persuasive.

Recently, agent simulations powered by Large Language Models (LLMs) have gained significant attention. Researchers have implemented these simulations at both individual and task levels, focusing on generating highly reliable human-like behavior~\cite{shao2023character,wang2023rolellm,xie2024can,qi2024representation} and facilitating multi-agent collaboration~\cite{du2023improving,hong2023metagpt,qian2024chatdev}. However, existing studies struggle to address massive population election simulations, as achieving the required levels of diversity and quantity remains difficult.

The U.S. Presidential Election, as a big election event, plays a pivotal role in shaping public engagement and party strategies~\cite{bartels1996uninformed,rosenstone1981forecasting}.
We use the U.S. presidential election as a case to explore effective methods for achieving massive and diverse election simulations with LLMs. To facilitate large-scale election simulations using LLMs, three primary challenges must be addressed.

\textbf{Q1. How to achieve high accuracy in individual-level simulations?}\\
The macro result is aggregated from individuals so that the preciseness of the individual directly impacts the overall outcome of the massive simulation. LLMs often lack sufficient personalized input for simulating the nuanced behavior of individuals. The limited contextual data restricts their ability to fully capture the diversity of voter behavior, motivations, and decision-making processes, which are critical for generating accurate and meaningful simulations.

\textbf{Q2. How to generate customized distributions that align with real-world statistics?}\\
Accurate election simulation requires that the simulated individuals represent the diversity and aligned distribution of real-world populations. While random sampling is able to capture this diversity, it falls short when aligning to the demographic distribution of the real world and is prone to source-driven biases~\cite{giorgi2022correcting, vraga2016party,cinelli2021echo, yusuf2014social, ribeiro2018media}. As a result, a carefully designed sampling strategy that mirrors real-world demographic and behavioral distributions is essential for producing valid and reliable simulations.

\textbf{Q3. How to evaluate the performance of election simulation in a systematic way?}\\
Evaluation metrics for election simulations vary depending on the specific context and task. Most existing works primarily focus on prediction accuracy, which offers a limited and unsystematic approach to assessing the full scope of simulation outcomes. Consequently, it is crucial to design a multi-aspects evaluation method to benchmark election results and provide comprehensive analyses.

We address these challenges by introducing \textbf{ElectionSim}: a massive population election simulation framework powered by large language models. For accurate individual simulation, We collect 171,210,066 tweets from Twitter between January 1, 2020, and December 29, 2020, to construct a large and diverse voter pool with \textbf{million-level distinct users}. For real-world voter distribution alignment, we employ a demographic sampling strategy to align the distribution between sampled users and real-world voters. To provide a systematic evaluation for the election simulation, we construct a \textbf{P}oll-based \textbf{P}residential \textbf{E}lection (\textbf{PPE}) benchmark,  integrating three evaluation baselines.

Extensive experiments show that LLMs achieve true-to-life performance compared to actual election outcomes following our ElectionSim framework. In voter-level simulations, we achieve a Micro-F1 score of 0.812 for vote-related tasks. At the state level, we accurately predict the outcomes of the 2020 presidential election in \textbf{47/51} states, and our predictions match the actual result in \textbf{12/15} battleground states as shown in Figure~\ref{fig:head pic}.

To conclude, we make four major contributions:
\begin{itemize}[leftmargin=*]
    \item \textbf{ElectionSim}: a massive population election simulation framework, which allows for diverse election simulating scenarios with high confidence by employing a flexible, customized distribution sampling strategy to align with the real world.
    \item \textbf{Million-level voters pool}: a large and diverse voters pool to support massive population election simulation by collecting and combing data from social media platforms.
    \item \textbf{PPE}: a poll-based presidential election benchmark to validate our simulation results. Experiments and further analyses confirm the benchmark's robustness and demonstrate the effectiveness of our framework in U.S. presidential election simulations.
    \item \textbf{Interactive Simulation Demonstration}: a demonstration system that facilitates multi-round dialogue interactions with voters we simulated. The system allows voters to be selected in a variety of ways, including specifying attributes and specifying responses to specific questions.
\end{itemize}

 In the following sections, we explore the core contents of ElectionSim and PPE. In Section~\ref{sec:framework}, we detail the key elements of the ElectionSim framework. In Section~\ref{sec:president election}, we introduce the PPE benchmark. Detailed experiment results are displayed in Section~\ref{sec:evaluation} with further analysis discussed in Section~\ref{sec:analysis}. Additionally, we provide a visualization of the U.S. presidential election in Section~\ref {app:visualization}. Supplementary materials including additional simulation experiments, data processing details, prompt libraries, case studies, and questionnaires can be found from Appendix~\ref{app:res_2024} to Appendix~\ref{questionnaire} respectively.

\begin{figure}[t]
    \centering
    \includegraphics[width=\linewidth]{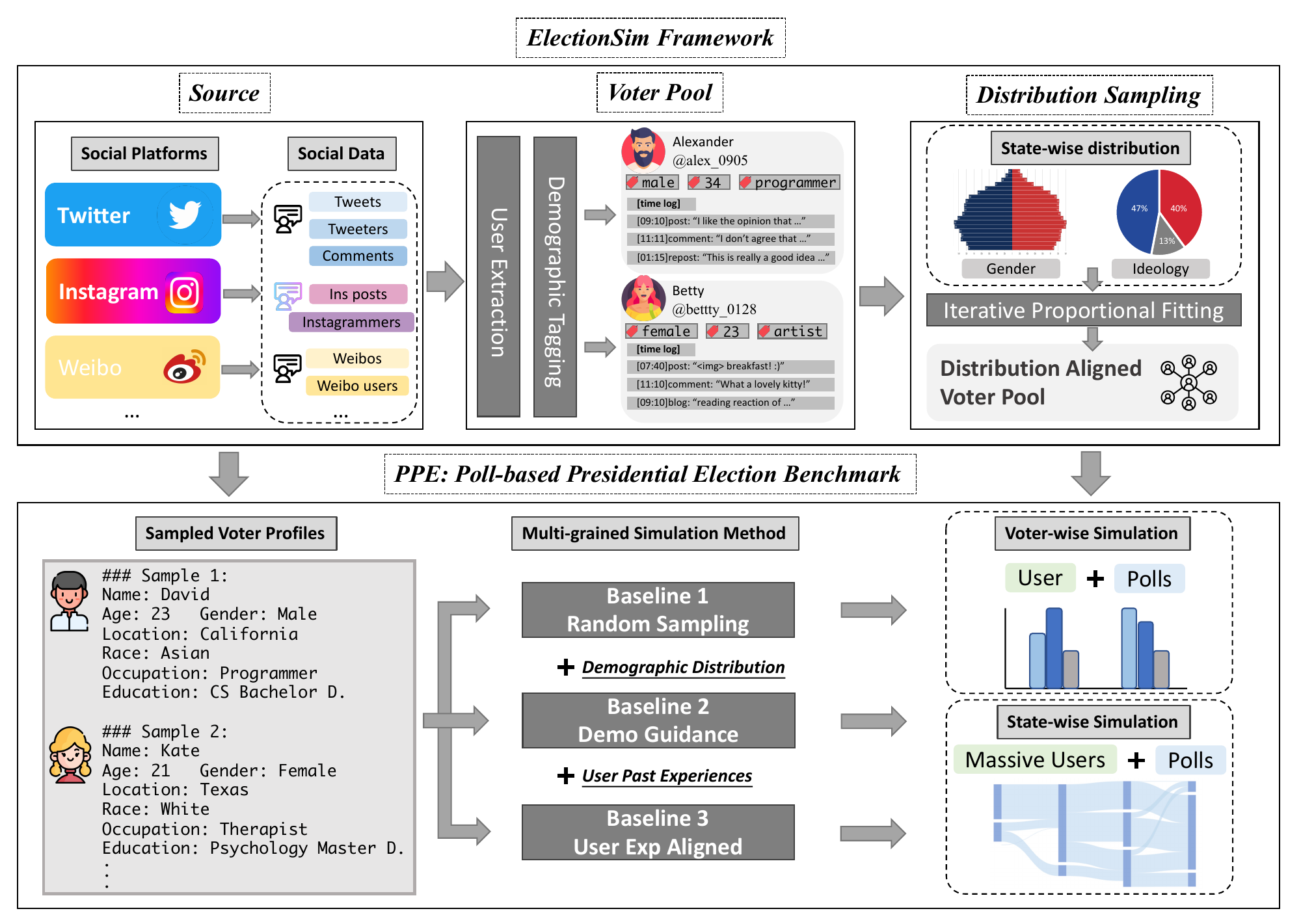}
    \caption{An illustration of the overall pipeline of the U.S. presidential election simulations.}
    \label{fig:overall pipeline}
\end{figure}

\section{ElectionSim: Massive Population Election Simulation}\label{sec:framework}
To address the challenges outlined in Section~\ref{sec:intro}, we introduce \textbf{ElectionSim}: a massive population election simulation framework. The overall pipeline is shown in Figure~\ref{fig:overall pipeline}.
The ElectionSim framework operates as follows: (1) Raw data, including user information and historical activities such as posts and comments, are collected from social platforms; (2) The data are processed on a per-user basis, and then a demographic tagging process, performed by specialized classifiers, generates a diverse and massive voter pool (\S\ref{sec:user pool}); (3) An interactive proportional fitting sampling strategy is used to approximate marginal distributions into a joint distribution for the target group sampling. After the pipeline of ElectionSim, we enable the sampling of any customized distribution from the voter pool, facilitating realistic massive election simulation (\S\ref{sec:sampling strategy}).

\subsection{Massive and Diverse Voter Pool Construction}\label{sec:user pool}
\subsubsection{Social Media Data Collection}\label{subsec:social media data}

\paragraph{Source Data} We choose Twitter\footnote{\url{https://x.com}} (now known as X)
as the primary data source due to its extensive user base. We collect tweets published by Twitter users in 2020, which allows us to construct a massive and varied voter pool that reflects diverse demographic and ideological perspectives. Table~\ref{tab:fields} details the specific fields we gather from the Twitter platform, and Table~\ref{tab:descriptive} shows the characteristics of the collected data.

\begin{table}[h!]
    \centering
    
    \label{tab:dataset}
    \begin{subtable}[b]{0.51\textwidth}
    \centering

    \adjustbox{max width=\textwidth}{
    \begin{tabular}{ccl}
\toprule[1.1pt]
\textbf{Dimension}   & \multicolumn{1}{c}{\textbf{Field}}  &  \multicolumn{1}{c}{\textbf{Description}} \\ \midrule
\multirow{2}{*}{User} & user\_id & The ID of the user who posted this tweet.\\
 & user\_at\_name & The name of the user who posted this tweet.\\ \hline
\multirow{4}{*}{Post} & tweet\_id & A unique ID to represent a tweet.\\
& tweet\_content & The content of this tweet.\\
& pub\_time & The publishing time of this tweet.\\
& lang & The language of this tweet. \\ \bottomrule[1.1pt]
\end{tabular}}
\caption{Data fields.}
\label{tab:fields}
\end{subtable}
\hfill 
\begin{subtable}[b]{0.45\textwidth}
    \centering

    \adjustbox{max width=\textwidth}{
   \begin{tabular}{lc} 
        \toprule[1.1pt]
        \textbf{Feature} & \textbf{Value} \\ 
        \midrule
        Number of Users & 9,596,198 \\
        Number of Tweets & 171,210,066 \\
        Avg. Number of Words in Tweets & 21.69 \\
        Time Span of Tweets & 2020.1.1--2020.12.29 \\
        Number of Languages in Tweets  & 50 \\
        \bottomrule[1.1pt]
    \end{tabular}}
    \caption{Descriptive statistics.}
    \label{tab:descriptive}
    \end{subtable}
    \caption{An overview of Twitter dataset characteristics.}
\end{table}

\paragraph{Data Preprocessing}\label{subsec:user extraction}
To tailor our voter pool for the context of the U.S. election, we implement a data-cleaning process consisting of four stages: user aggregation, language filtering, post filtering, and user cleaning.

\begin{figure}[h!]
    \centering
    \begin{subfigure}[b]{0.24\textwidth}
        \centering
        \includegraphics[width=\linewidth]{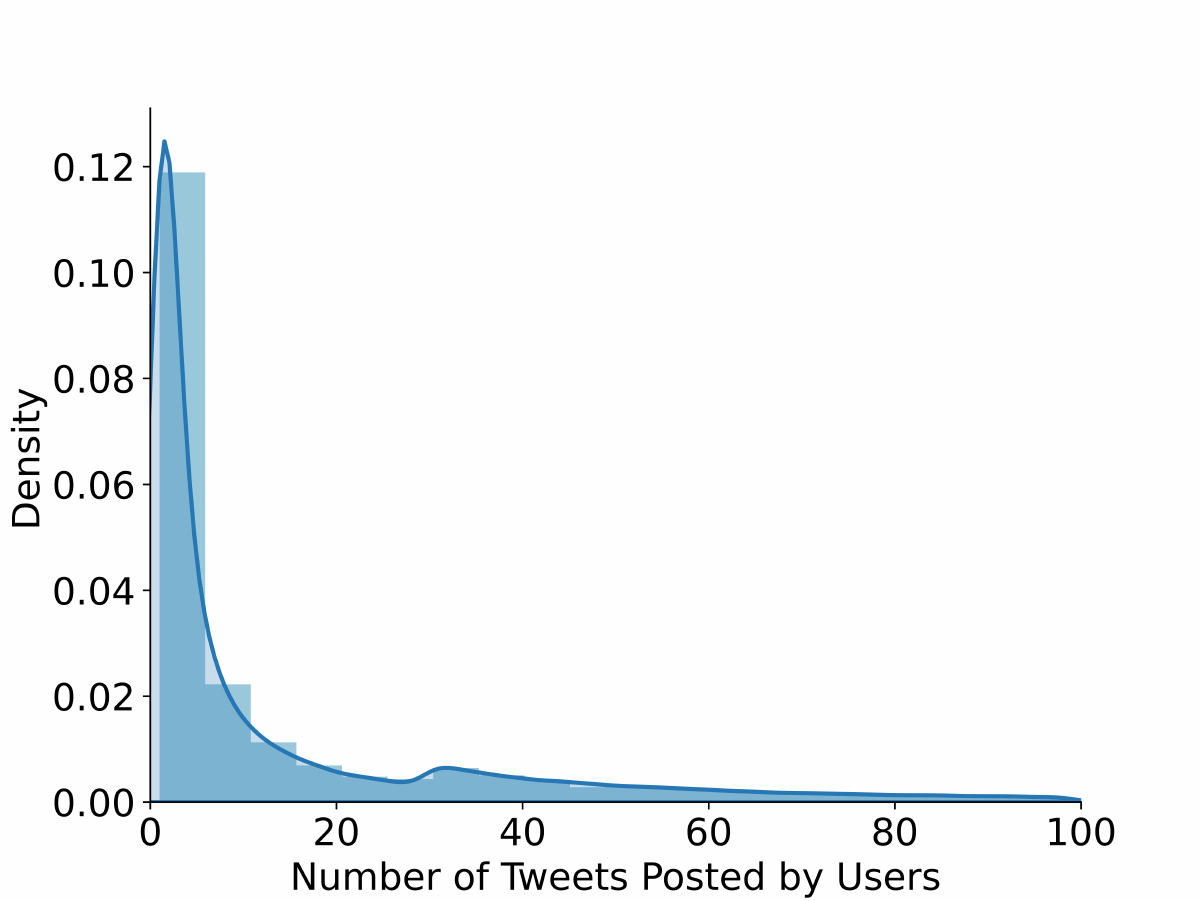}
        \caption{}
    \end{subfigure}
    \begin{subfigure}[b]{0.24\textwidth}
        \centering
        \includegraphics[width=\linewidth]{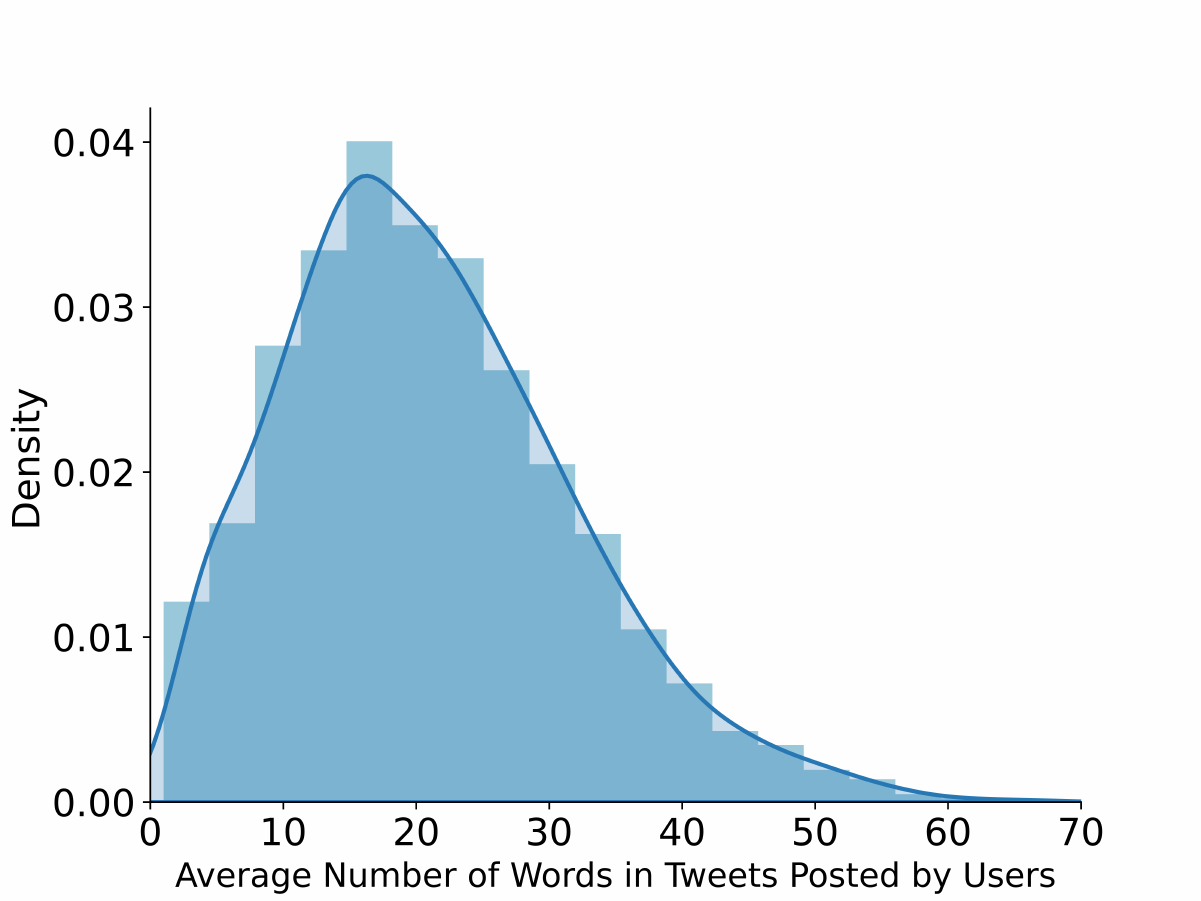}
        \caption{}
    \end{subfigure}
    \vspace{1em} 
    \begin{subfigure}[b]{0.24\textwidth}
        \centering
        \includegraphics[width=\linewidth]{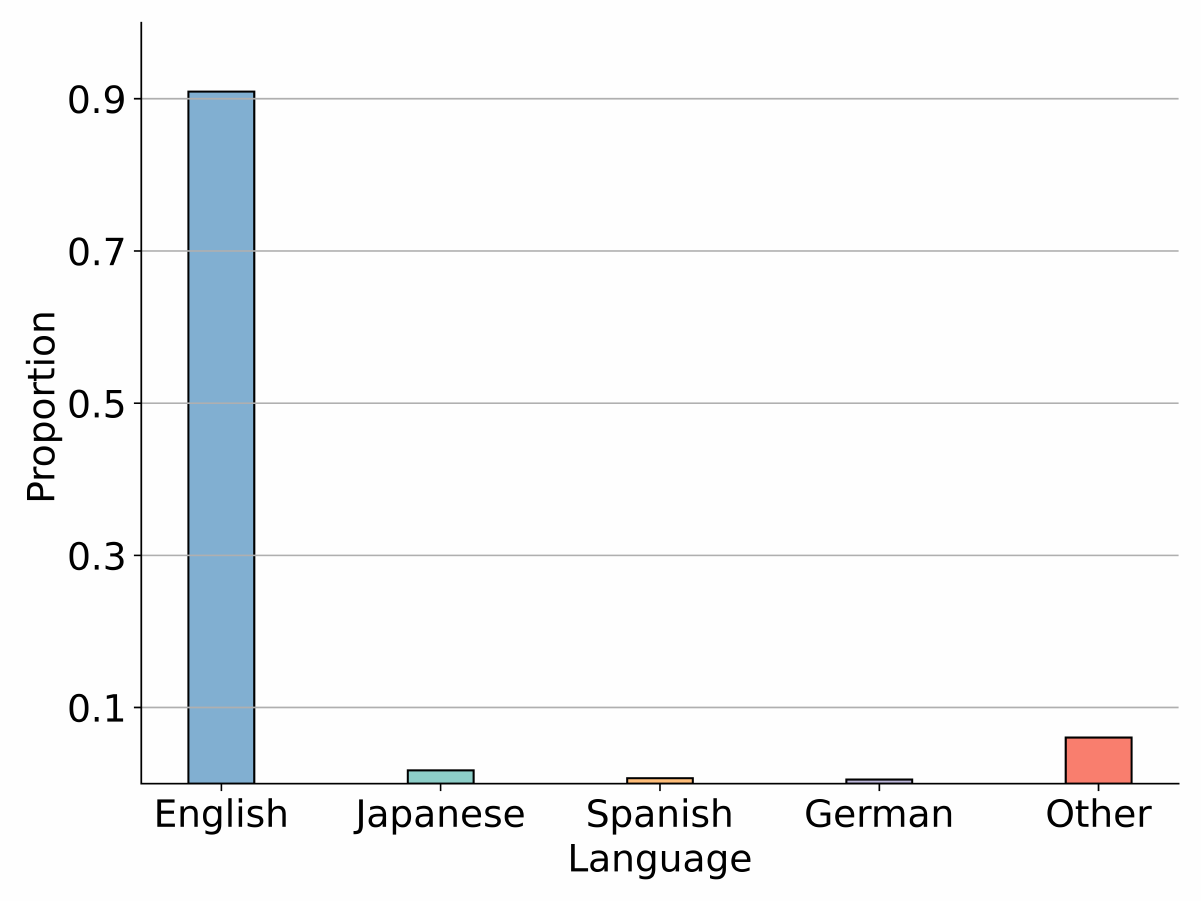} 
        \caption{}
    \end{subfigure}
    \begin{subfigure}[b]{0.24\textwidth}
        \centering
        \includegraphics[width=\linewidth]{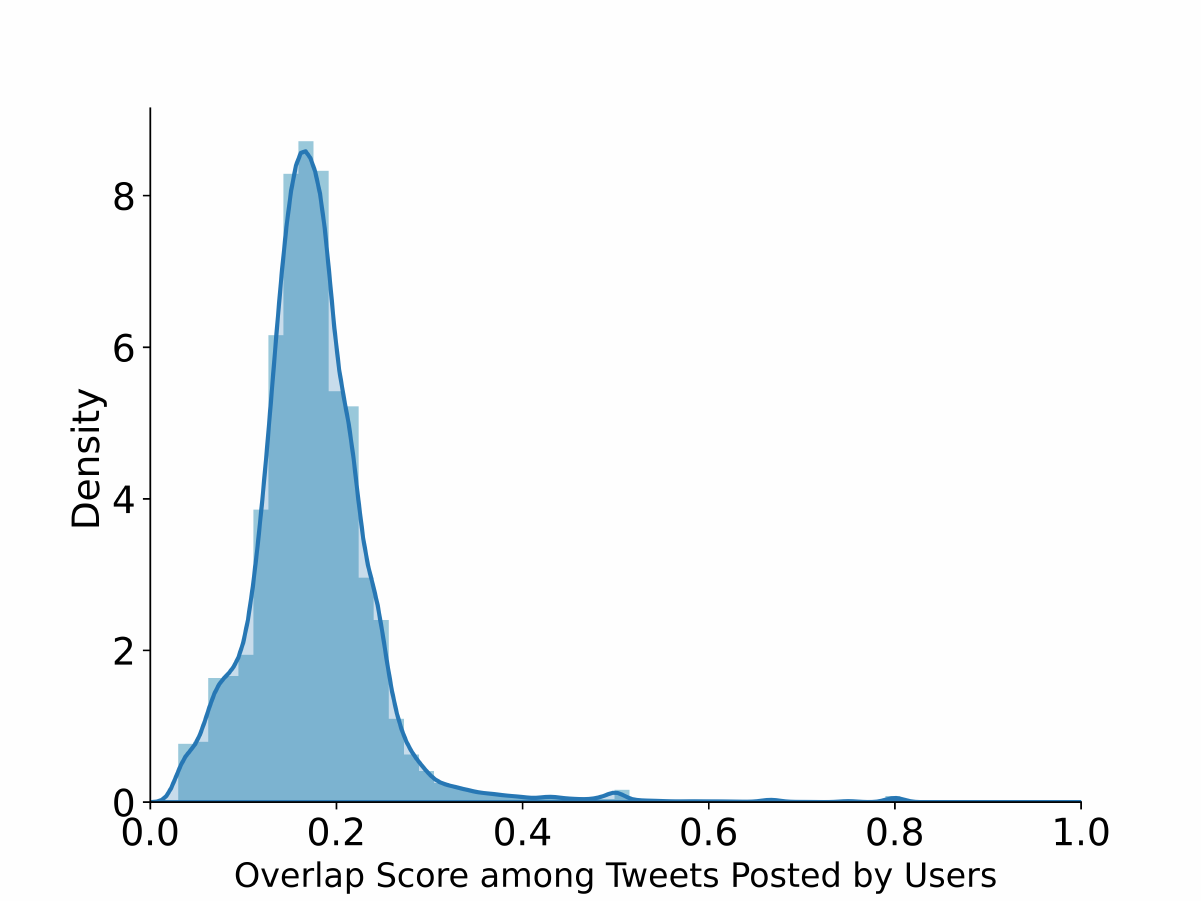} 
        \caption{}
    \end{subfigure}
    \caption{Statistical charts based on user aggregation, including:
a) Density plot of tweets per user,
b) Density plot of average word count per tweet,
c) User distribution across languages, and
d) Density plot of user overlap scores.
}
    \label{fig:statistics_user}
\end{figure}

\begin{itemize}[leftmargin=2em]
    \item \textbf{\emph{User Aggregation}} The data collected is initially structured at the post level; in order to construct a voter pool, we aggregate the data to a user-level framework. The key statistical results after this aggregation process are presented in Figure~\ref{fig:statistics_user}.
    \item \textbf{\emph{Language Filtering}}We start by removing non-English posts to better reflect the perspectives of American users (detailed in Appendix~\ref{app:filtering}). This is important for accurately interpreting sentiments related to the U.S. election, as language can significantly shape political discourse \cite{perez2017language, perez2019language}. 
    \item \textbf{\emph{Post Filtering}} Next, we retain only those users who have made over 30 posts and sample 30 of them from all posts as their historical information. 
    \item \textbf{\emph{User Cleaning}} Finally, we evaluate content repeatability by calculating the overlap score from a random sample of 5 tweets from each user. We compute the Jaccard scores for all pairs, and the final score is the average of these values (detailed in Appendix~\ref{app:jaccard}). 
\end{itemize}

\begin{table}[htbp]
    \centering
    \renewcommand{\arraystretch}{1.25} 
    \begin{tabular}{lc} 
        \toprule
        \textbf{Metric} & \textbf{Value} \\ 
        \midrule
        Number of Users & 1,006,517 \\
        Number of Tweets & 30,195,510 \\
        Avg. Number of Words in Tweets & 22.36 \\
        \bottomrule
    \end{tabular}
    \caption{Statistical summary of the processed Twitter user dataset.} 
    \label{tab:user_summary}
\end{table}

By applying these stages, we enhance the reliability and relevance of our dataset, making it more suitable for analyzing user behaviors in the context of the U.S. election. Table~\ref{tab:user_summary} presents the statistical summary of the processed Twitter user data.

\subsubsection{Demographic Feature Annotation}\label{subsec:demographic annotation}
The data collected from Twitter primarily consists of user post histories and lacks essential demographic features necessary for effective social group simulation. To enhance the dataset with reliable demographic attributes, we develop a comprehensive taxonomy of demographic features, accompanied by corresponding classifiers to annotate the Twitter data systematically. 

\paragraph{Demographic Taxonomy}
We conduct an analysis of multiple U.S. polls~\cite{anes2021,bartels1999panel, miller1989american} and synthesize the demographic classifications pertinent to these studies. Through this process, we develop a comprehensive demographic taxonomy that is closely aligned with the context of the U.S. election, as shown in Table~\ref{tab:taxonomy}.

\begin{table}[h]
\centering
\begin{tabular}{cccc}
\toprule
Dimension & Attribute & Classification \\
\midrule
\multirow{9}{*}{Personal Traits} & \multirow{2}{*}{Gender} & Male\\
& & Female\\
\cmidrule(lr){2-3}
 & \multirow{3}{*}{Age} & Youth (18-35 years old) \\ 
 & & Middle-aged (36-65 years old) \\
 & & Elderly (over 65 years old) \\
\cmidrule(lr){2-3}
 & \multirow{4}{*}{Race} & White \\
   & & Black \\
    & & Asian \\
     & & Hispanic \\
\midrule
\multirow{7}{*}{Political Orientation} & \multirow{3}{*}{Ideology} & Liberal \\
     & & Moderate \\
    & & Conservative \\
\cmidrule(lr){2-3}
 & \multirow{4}{*}{Partisanship} & Democrat \\
     & & Republican \\
    & & Independent \\
    & & Others \\
\bottomrule
\end{tabular}
\caption{The demographic label taxonomy.}
\label{tab:taxonomy}
\end{table}

\paragraph{Demographic Classifier}

To develop a demographic dataset for classifiers, we must establish a ground-truth test dataset for evaluation and a reliable training dataset for effective model training. The following outlines the process for constructing these datasets and training the classifiers.

\begin{itemize}[leftmargin=*]
    \item \textbf{\emph{Test Set Construction}} We use a semi-automated method to create the ground-truth test data in two stages: API annotation and manual verification. First, we sample 200 users from the Twitter dataset and annotate them using advanced commercial LLM APIs based on our demographic taxonomy (detailed in Appendix~\ref{app:test_building}), including GPT-4o\footnote{\texttt{gpt-4o-2024-08-06}}, Claude-3.5\footnote{\texttt{claude-3-5-sonnet-20240620}}, and Gemini-1.5\footnote{\texttt{gemini-1.5-pro}}. We then engage 5 professional annotators to validate the API-generated annotations and finalize the ground-truth labels.

    \item \textbf{\emph{Train Set Construction}} To create a large train dataset, we utilize commercial APIs for annotation. We evaluate the consistency between each API's annotations and the ground-truth labels on the test set, as well as the consistency between the majority voting results from these APIs and the ground-truth labels (detailed in Appendix~\ref{app:consistency}, results shown in Figure~\ref{fig:consistency}). It shows that majority voting achieves the highest consistency. Consequently, we sample 10,000 users from the processed Twitter dataset, label them using three APIs, and adopt the majority voting results as the labels for our training set.

    \item \textbf{\emph{Implementation Details}} 
We choose the Longformer\footnote{\url{https://huggingface.co/allenai/longformer-base-4096}}
~\cite{beltagy2020longformer} model as the backbone for our classifiers due to its capability to process long text windows of up to 4096 tokens. We develop classifiers for each demographic feature, resulting in a total of 5 classifiers. For each classifier, we implement full parameter fine-tuning, set the learning rate $5 \times 10^{-5}$, batch size to 16, employ AdamW as the optimizer, and train the model 3 epochs on 8 NVIDIA RTX4090 GPUs. 
    \item \textbf{\emph{Classifiers Performance}} For each demographic feature, we select the checkpoint with the highest F1 score on the test set as the final model. The performance of the classifiers obtained on the test set is shown in Figure~\ref{fig:classifier_performance}.

\end{itemize}

\begin{figure}[h!]
    \begin{subfigure}[t]{0.49\textwidth}
        \centering
        \includegraphics[width=\linewidth]{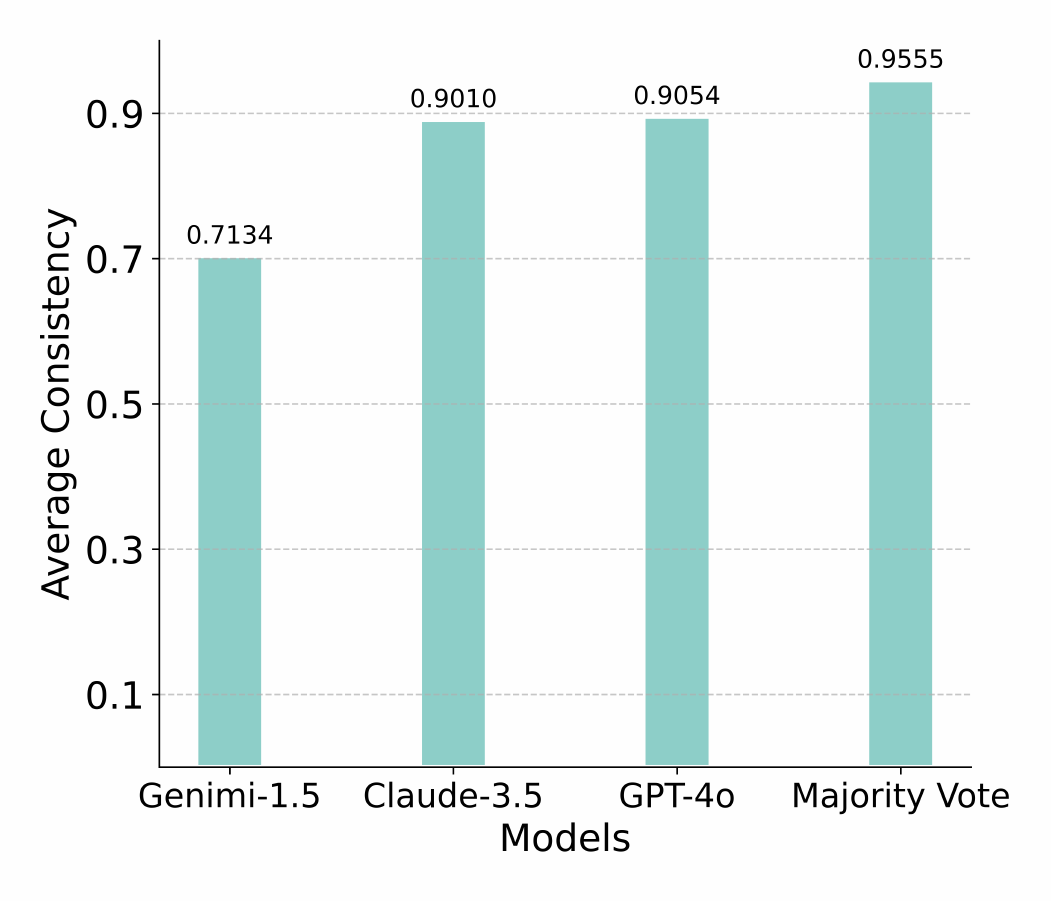}
        \caption{Consistency between commercial APIs and manual annotations on the test set}
        \label{fig:consistency}
    \end{subfigure}
    \begin{subfigure}[t]{0.48\textwidth}
        \centering
        \includegraphics[width=\linewidth]{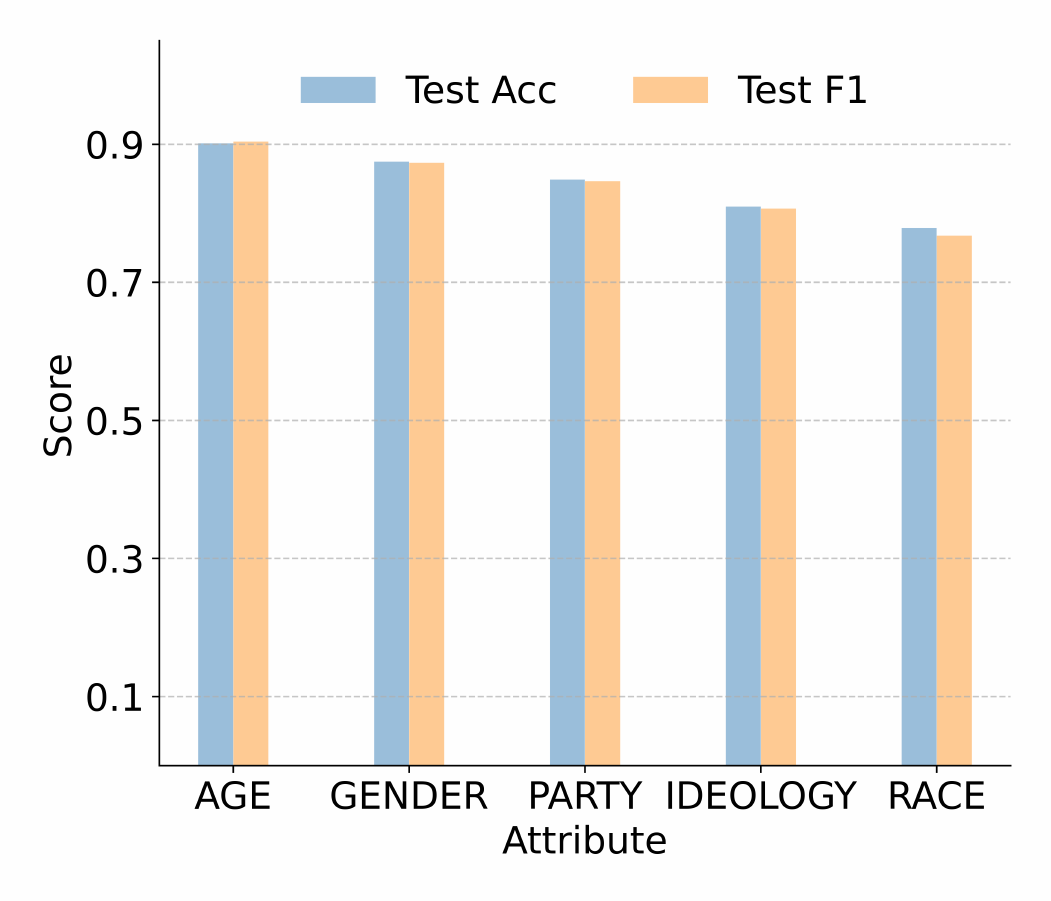}
        \caption{The accuracy and F1 score of each classifier on the test set}
        \label{fig:classifier_performance}
    \end{subfigure}
    \caption{Classification accuracy of API-based LLMs and our demographic classifiers.}
\end{figure}

\begin{figure}[h!]
    \centering
    \includegraphics[width=0.6\linewidth]{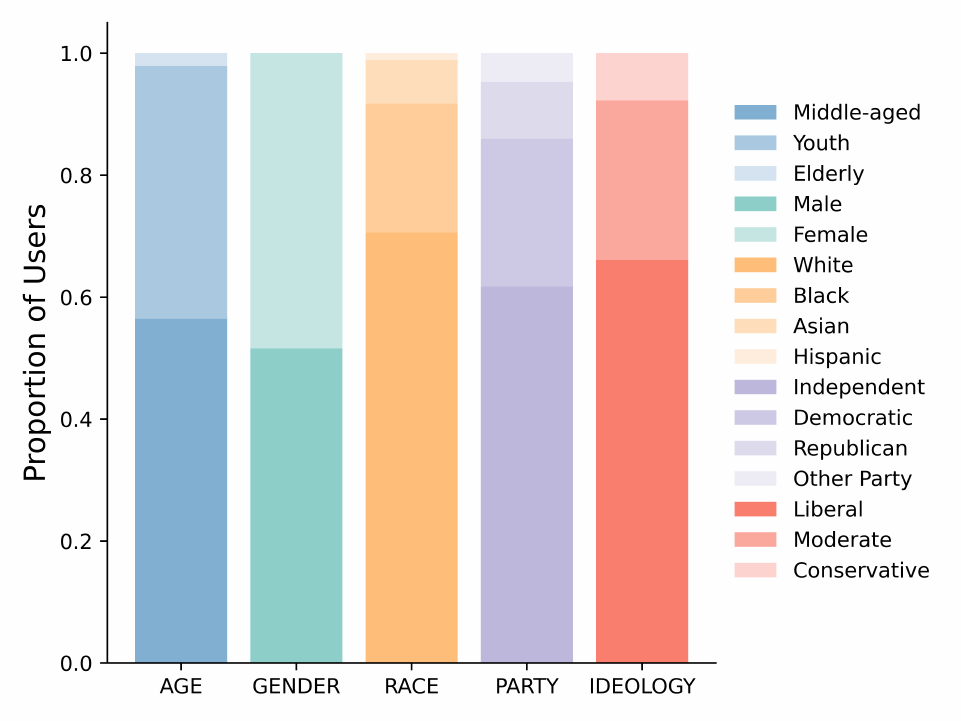}
    \caption{Distribution of attribute categories in the voter pool}
    \label{fig:annotation}
\end{figure}

\subsection{Real-world Demographic Distribution Sampling Strategy}\label{sec:sampling strategy}
\subsubsection{Real-World Demographics in Electoral Processes}

Demographics play a significant role in U.S. elections, including presidential races. Political science literature widely acknowledges that factors such as gender, race, and income level influence U.S. citizens' voting decisions~\cite{major2018threat, teixeira2009red}. Moreover, evolving demographic dynamics are significantly impacting candidates' campaign strategies and the overall election outcomes. For instance, as the number of Hispanic voters grows, many are shifting away from the Democratic Party and becoming more receptive to Republican candidates~\cite{NYT}, which is notably affecting electoral results in states with large Hispanic populations, such as Texas and Florida. In addition to demographics, ideology and partisanship are also critical predictors of U.S. voting behavior. Therefore, to ensure our large-scale simulation of election results closely mirrors reality, it is essential to incorporate accurate distributions of demographics, ideology, and partisanship. In the remainder of this section, we introduce the methods used to simulate the distribution of these real-world attributes.

\subsubsection{Methods for Distribution Sampling}

To accurately reflect the demographic and ideological makeup of U.S. citizens, we combine multiple datasets to construct the population distribution in our study. Specifically, we utilize data from the U.S. Census Bureau’s Voting and Registration in the Election of November 2022, along with the 2020 Time Series Study from the American National Election Studies (ANES)~\cite{anes2021}. The U.S. Census Bureau has been collecting data on the characteristics of American voters since 1964. This dataset primarily tracks how many citizens of voting age are registered and how many vote, broken down by factors such as age, gender, race, ethnicity, and more. In contrast, the American National Election Studies (ANES) consist of academically-run national surveys of U.S. voters, conducted before and after every presidential election. The ANES data extend a long tradition of studies, with records dating back to 1948. The two datasets allow us to simulate real population distributions based on variables such as state of registration, gender, age, race, ideology, and partisanship. 

The datasets chosen are the most recent and frequently cited in U.S. election studies~\cite{bartels1999panel, miller1989american}, while the selected features also capture fundamental demographic and ideological attributes that are key predictors of voting behavior~\cite{wang2015forecasting}. Table~\ref{tab: distribution_sample} provides the sources of all the attributes used in the population simulation, along with the classifications for each attribute.

\begin{table}[h]
\centering
\begin{tabular}{cccc}
\toprule
Dataset & Attribute & Classification \\
\midrule
\multirow{11}{*}{U.S. Census Bureau} & \multirow{2}{*}{Gender} & Male\\
& & Female\\
\cmidrule(lr){2-3}
 & \multirow{5}{*}{Age} & 18-24 years old \\ 
 & & 25-34 years old \\
 & & 35-44 years old \\
 & & 45-64 years old\\ 
 & & over 65 years old \\
\cmidrule(lr){2-3}
 & \multirow{4}{*}{Race} & White \\
   & & Black \\
    & & Asian \\
     & & Hispanic \\
\midrule
\multirow{7}{*}{ANES} & \multirow{3}{*}{Ideology} & Liberal \\
     & & Moderate \\
    & & Conservative \\
\cmidrule(lr){2-3}
 & \multirow{4}{*}{Partisanship} & Democrat \\
     & & Republican \\
    & & Independent \\
    & & Others \\
\bottomrule
\end{tabular}
\caption{The Demographic Distribution Sampling Attributes and Classifications}
\label{tab: distribution_sample}
\end{table}

The U.S. Census Bureau provides accurate demographic data on voters across all 50 states and the District of Columbia for 2022, including variables such as gender, race, and age, which we directly incorporate into our simulation. In contrast, the ANES dataset samples voters from each state and includes individual interview responses. Therefore, the dataset does not provide direct information about the distribution of voters' ideology, partisanship, or community type at the state level. However, according to the ANES methodology report~\cite{debell2022methodology}, the sampling strategy involves random selection based on mailing addresses and is stratified by area. This method produces a sample that closely mirrors the true demographic and ideological distribution of U.S. citizens across states. As a result, we treat the distribution of these attributes in the ANES data as representative of their actual distribution in each state.

\subsubsection{Iterative Proportional Fitting}

In addition to using the distributions of individual variables to capture population characteristics, we also generate the joint distribution of all variables to more accurately represent voters in each state. Specifically, we apply Iterative Proportional Fitting (IPF) to estimate the joint distribution of all attributes within each state. IPF is an iterative method that estimates missing values for joint population attributes, providing the foundation for a more comprehensive simulation. This method is widely used in social science and population research~\cite{choupani2016population}.

In our study, we follow the classical IPF method to construct the joint distribution of all the attributes in our simulation. Specifically, we start with a two-way table with individual components denoted as $x_{ij}$ and targeted estimation $\hat{x}_{ij}$. The targeted estimation $\hat{x}_{ij}$ satisfies $\Sigma_j \hat{x}_{ij} = v_i$ and $\Sigma_i \hat{x}_{ij} = w_j$. The iterations are specified as follows:

\begin{center}
\begin{minipage}{0.8\textwidth} 
Let $\hat{x}_{ij}^{(0)} = x_{ij}$. For $\alpha > 1$: 

\begin{equation*}
 \hat{x}_{ij}^{(2\alpha-1)} = \dfrac{\hat{x}_{ij}^{(2\alpha - 2)}v_i}{\Sigma^J_{k=1}\hat{x}_{ij}^{(2\alpha - 2)}}   
\end{equation*}

\begin{equation*}
\hat{x}_{ij}^{(2\alpha)} = \dfrac{\hat{x}_{ij}^{(2\alpha - 1)}w_j}{\Sigma^I_{k=1}\hat{x}_{ij}^{(2\alpha - 1)}}  
\end{equation*}

\end{minipage}
\end{center}

The iterations end when the estimated marginals are sufficiently close to the real marginals or when they stabilize without further convergence.

In our study, we implement the IPF algorithm for each state using five attributes: gender, race, age group, ideology, and partisanship. In most cases, the algorithm does not converge, but the gaps between the estimated and actual marginals are less than 5\%, with 888 out of 918 marginals falling within this range. For the outliers, since IPF adjusts proportionally to the marginals, the overall ratio of marginals remains consistent. We then use the estimated joint distribution and marginals for our massive simulation.

\section{PPE: Poll-based Presidential Election Benchmark}\label{sec:president election}
In this section, we introduce the U.S. presidential election prediction benchmark via a poll-based survey method, namely \textbf{PPE}. The task is clarified in \S\ref{subsec:president task} and the source of the benchmark is introduced in \S\ref{subsec:poll and census}. \S\ref{subsec:questionnaire} provides a comprehensive explanation of the questionnaire design process, along with an overview of the questionnaire's statistics. Furthermore, we introduce the multi-stage evaluation baseline in \S\ref{subsec:eval_baseline}, enabling the massive simulation execution.

\subsection{Task Definition}\label{subsec:president task}
The poll-based presidential election prediction task is an application example of the massive social simulation paradigm. This task simulates the poll results of each person within a group at a fine granularity and aggregates the results according to certain distribution patterns (such as demographic information at the state level) to obtain the macro-level social opinion tendencies.

\subsection{Poll Collection}\label{subsec:poll and census}
We use the American National Election Studies (ANES) 2020 Time Series Study data as the source of the benchmark~\cite{anes2021}. The ANES 2020 Time Series Study is a continuation of a series of studies conducted since 1948 to enable analysis of public opinion and voting behavior in U.S. presidential elections. The 2020 study features a two-wave panel design with pre-election and post-election interviews.

The ANES 2020 questionnaires consist of two parts: the Screener Questionnaire and the Survey Questionnaire. The Survey Questionnaire is further divided into pre-election and post-election sections. It primarily consists of multiple-choice questions, with a few multiple-answer and fill-in-the-blank questions. The questionnaire covers a broad range of topics, with the approximate percentage of each topic indicated in Table ~\ref{tab:percentage}.

\begin{table}[htbp]
    \centering
    \renewcommand{\arraystretch}{1.25}
    \begin{tabular}{cc}
        \toprule
        \textbf{Percentage} & \textbf{Topic} \\
        \midrule
        10\% & Voting behavior \\
        7\% & Candidate evaluations \\
        3\% & Party evaluations \\
        12\% & Evaluations of government and politics \\
        13\% & Demographics \\
        7\% & Personal experience and outlook \\
        9\% & Political engagement \\
        4\% & Predispositions (traits, values, etc.) \\
        13\% & Group identities and attitudes \\
        19\% & Political issues \\
        3\% & Other \\
        \bottomrule
    \end{tabular}
    \caption{Approximate percentage of each topic in 2020 ANES survey questionnaire}
    \label{tab:percentage}
\end{table}

\begin{figure}[htbp]
    \centering
    \includegraphics[width=0.7\linewidth]{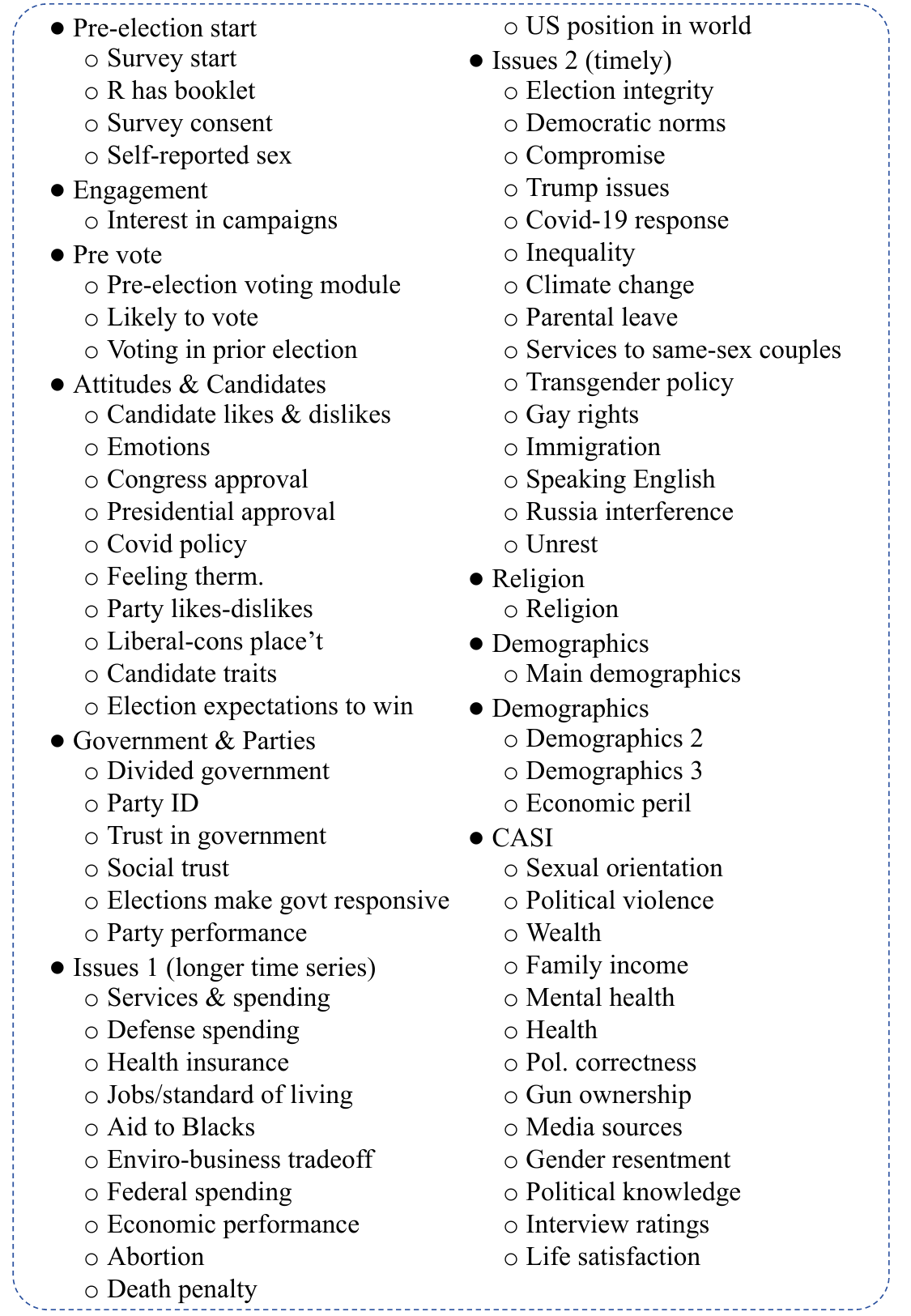}
    \caption{Pre-election questionnaire modules and sections: ANES 2020 Time Series Study}
    \label{questionnaire modules}
\end{figure}

The questionnaire is organized in modules, with each cohering on one or more dimensions of format, mode, or subject matter. These modules are further subdivided into sections and questions. Figure~\ref{questionnaire modules} outlines the modules and their component sections for the 2020 pre-election study.

A total of 8,280 respondents participated in the pre-election interviews, of whom 7,453 also completed the post-election reinterviews, while 827 did not respond. Most questions in the questionnaire are logically related, and depending on the respondents' answers, follow-up questions are asked accordingly. As a result, the number of questions asked varied for each respondent. Table ~\ref{tab:ANES_questionnaire_summary} provides a summary of the 2020 ANES survey questionnaire.

\begin{table}[htbp]
    \centering
    \renewcommand{\arraystretch}{1.25}
    \begin{tabular}{ccc}
        \toprule
        \textbf{Metric} & \multicolumn{2}{c}{\textbf{Value}}\\
        \midrule
         & Pre-election & Post-election \\\cline{2-3}
        Number of Questions& 443 & 511\\
        Number of Respondents & 8280 & 7453\\
        Time Span & Aug 18, 2020 - Nov 2, 2020 & Nov 8, 2020 - Jan 4, 2021\\
        \makecell{Avg. Number of Questions \\Answered per Respondent} & 341.68 & 349.83\\
        \bottomrule
    \end{tabular}
    \caption{2020 ANES survey questionnaire summary.}
    \label{tab:ANES_questionnaire_summary}
\end{table}

\subsection{Poll-based Questionnaire Design}\label{subsec:questionnaire}
The questionnaire used for the massive social simulation is modified based on the ANES data to better suit the needs of simulating real voters through social agents. Using the 2020 ANES questionnaire as a prototype, we design our questionnaire following these steps:

\paragraph{Selecting Topics} 
  We select a total of 24 socially significant topics and collect questions that reflect respondents' opinions on these topics. Each topic is associated with a varying number of specific questions, ranging from 1 to 6. Table~\ref{tab:questionnaire issues} presents a list of the topics and the number of questions corresponding to each.

\begin{table}[htbp]
    \centering
    \renewcommand{\arraystretch}{1.25} 
    \begin{tabular}{cc} 
        \toprule
        \textbf{Number of questions} & \textbf{Topic} \\ 
        \midrule
        $\geq 5$ & Democratic Norms, Immigration, LGBTQ+ Rights \\
        4 & Environment, Government \\
        2 & \makecell{Criminal Justice, Education, Gender Resentment, \\Health Care, Social Welfare, US Position in World} \\
        1 & \makecell{Abortion, Aid to Blacks, Aid to Poor, Defense, Economy, \\Election Integrity, Inequality, Infrastructure, Parental Leave, \\Social Security, Taxes, Unrest, Voting Behavior}\\
        \bottomrule
    \end{tabular}
    \caption{Selected questionnaire topics.} 
    \label{tab:questionnaire issues}
\end{table}

\paragraph{Optimizing Question Selection}
To reduce the complexity of the questionnaire while retaining as much original information as possible, we implement the following steps:

\begin{enumerate}[leftmargin=*]
    \item Removing multiple-answer questions and fill-in-the-blank questions.
    \item Eliminating conditional questions, i.e., those asked only to respondents who choose specific options in previous questions, ensuring that the remaining questions apply to all respondents.
\end{enumerate}

\paragraph{Merging Invalid Options} 
We define invalid options as those like "Refused", "Don’t know", or "Haven’t thought much about this", which do not provide clear responses. Having too many such options in the questionnaire may interfere with the agent’s performance in answering questions. We merge these invalid options into a single new category: "DK/RF."

\paragraph{Converting Intensity Questions to Position Questions} 

\begin{figure}[htbp]
    \centering
    \includegraphics[width=\linewidth]{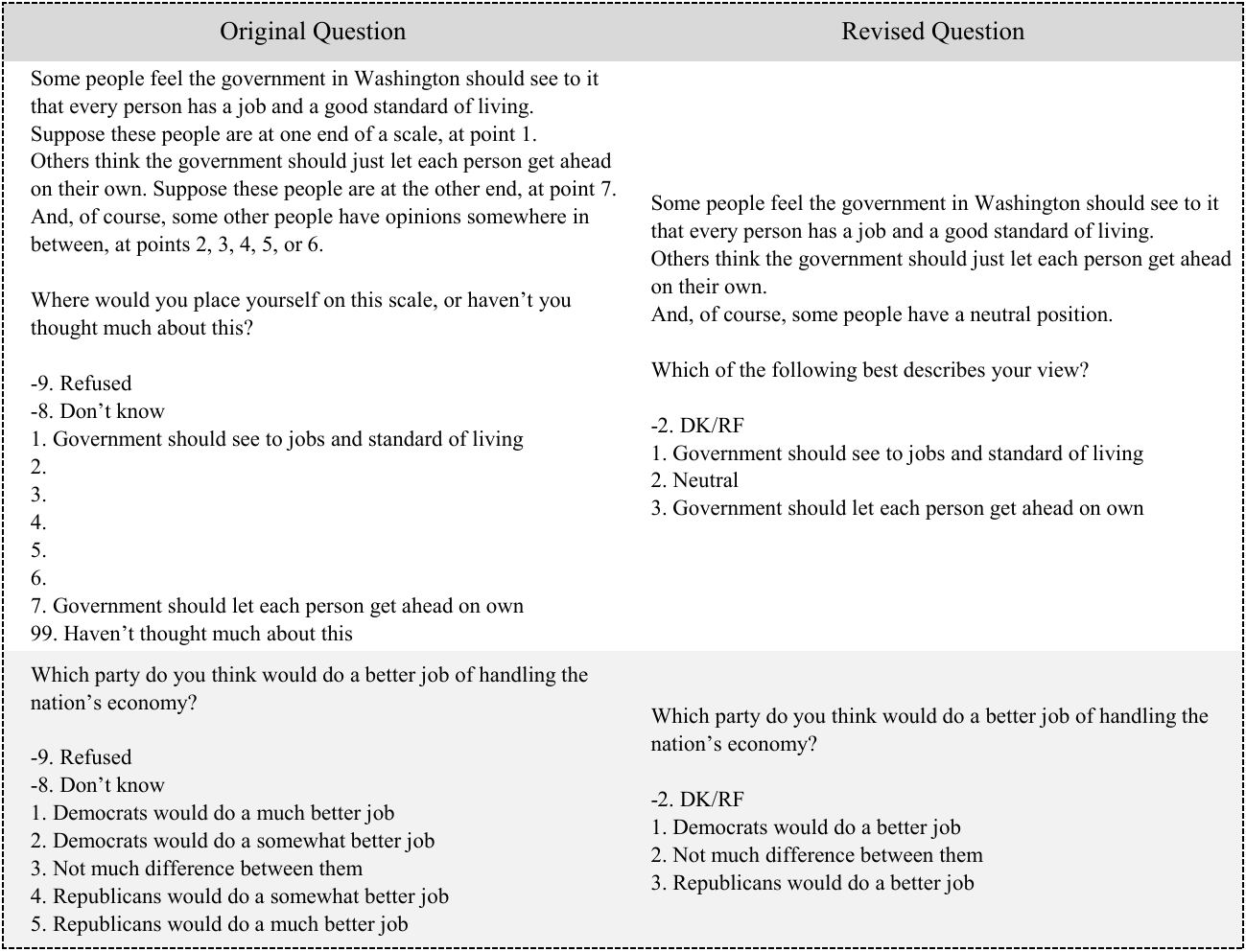}
    \caption{Revised Question Example.} 
    \label{Example: revised questions}
\end{figure}

In the original ANES questionnaire, some questions are overly detailed, asking respondents to specify the intensity of their stance, such as distinguishing between "Agree strongly" and "Agree somewhat." This level of detail imposes overly high demands on agents to align with human opinions. In redesigning the questionnaire, we ignore the intensity of opinions and only ask about the respondents' basic stance. We also convert the original 7-point Likert scale to a 3-point scale. This reduction in the number of options makes it easier for the agents to make choices. Figure~\ref{Example: revised questions} gives two examples of revised questions.

In the end, we finalize 49 questions across 24 issues. Table ~\ref{tab:questionnaire_summary} gives some general information about the questionnaire we designed. Most questions have 3 to 5 options, allowing for clear distinctions between different positions without making the choices too ambiguous. The full questionnaire can be found in Appendix ~\ref{questionnaire}.

\begin{table}[htbp]
    \centering
    \renewcommand{\arraystretch}{1.25} 
    \begin{tabular}{lc} 
        \toprule
        \textbf{Metric} & \textbf{Value} \\ 
        \midrule
        Number of Questions& 49 \\
        Number of Topics & 24 \\
        Avg. Number of Words per Question & 34.06 \\
        Avg. Number of Options per Question & 3.22 \\
        Number of Respondents & 8280 \\
        \bottomrule
    \end{tabular}
    \caption{Questionnaire Summary.} 
    \label{tab:questionnaire_summary}
\end{table}

\begin{figure}[t!]
    \centering
    \includegraphics[width=\linewidth]{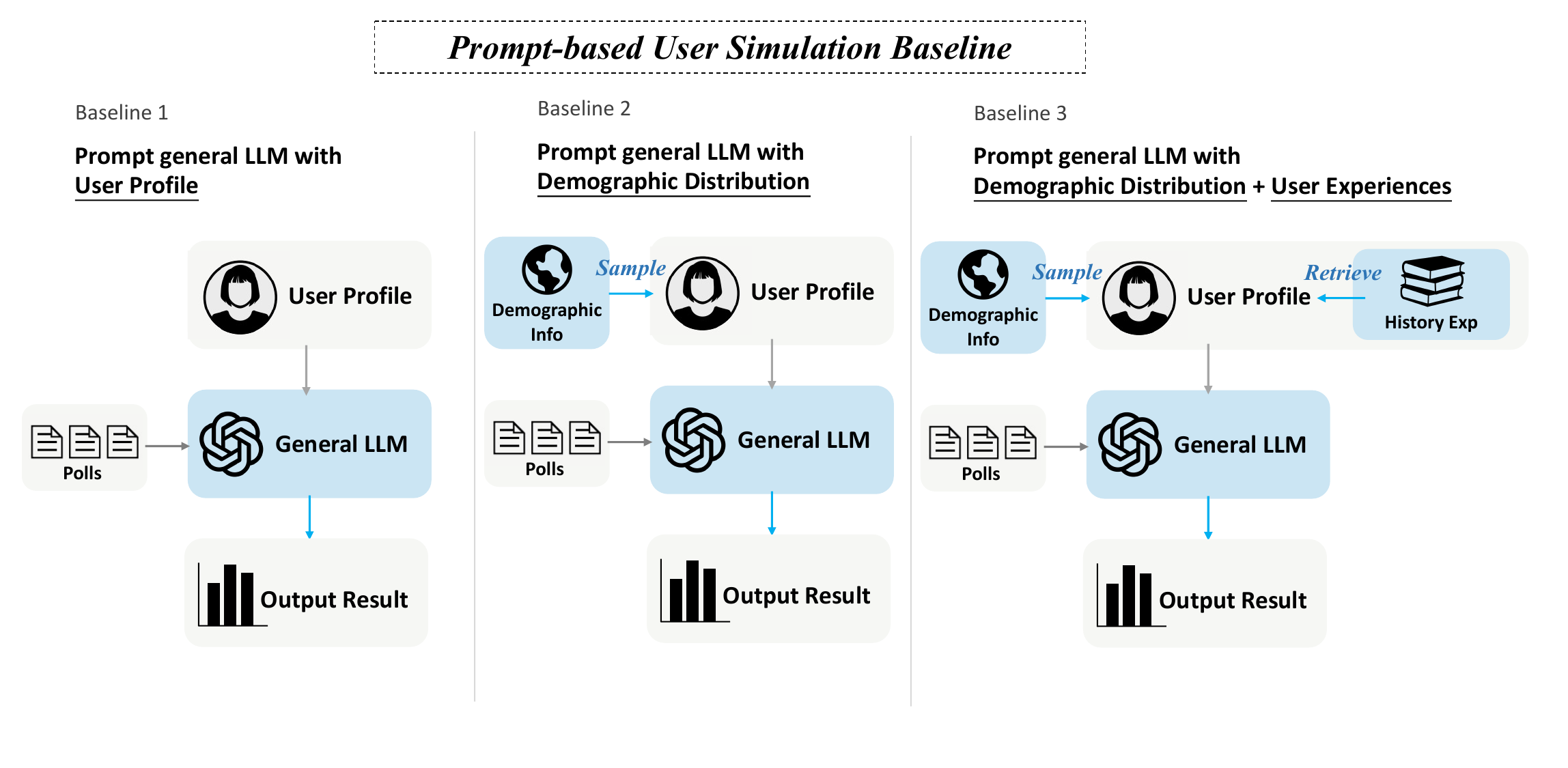}
    \caption{An illustration of prompt-based evaluation baselines.}
    \label{fig:baselines}
\end{figure}

\subsection{Prompt-based Evaluation Baselines}\label{subsec:eval_baseline}
We design three prompt-based evaluation baselines to evaluate the simulation results comprehensively, as shown in Figure~\ref{fig:baselines}.

\paragraph{Baseline 1: Prompt LLM with user profile} After the user pool construction, one simple way to carry out the social simulation is to randomly sample users from the user pool and interview them with the poll question. The overall distribution of the poll answers can partially reflect the group's preference towards different candidates and topics.

\paragraph{Baseline 2: Prompt LLM with demographic distribution} To better impend over the real-world results, we add demographic distribution information during the user preparation. We sample the users according to the census data according to the sampling strategy mentioned in~\S\ref{sec:sampling strategy}, which includes the demographic distribution of the voters across different states. The same poll-based interview task is carried out after the sampling.

\paragraph{Baseline 3: Prompt LLM with multiple information} Though demographic information adds diversity to the group sampling. Different individuals with similar demographic features are hard to distinguish. Baseline 2 may sample randomly from a subset of the user pool, the individuals within which are all "Middle-income graduated Asian males in their 20s". Thus, we add specific users' experiences, namely posts on social media, to the sampled user profiles to make each individual lively.

\section{Evaluation}\label{sec:evaluation}
In this section, we display the experiment results carried out under the PPE benchmark at both~\S\ref{subsec:voter exp} voter level and~\S\ref{subsec:state exp} state level.

\subsection{Voter-wise Simulation}\label{subsec:voter exp}
The first step of Poll-based President Election Prediction is to conduct fine-grained modeling of each voter's opinion behavior. Therefore, to verify the accuracy of the modeling for an individual, we conduct a voter-wise simulation. The task can be described as: given a series of demographic tag information of an individual, predict his/her response to a specific question.

\paragraph{Model} We select a series of models for performance comparison, which include both open-source and commercial large language models:

\begin{itemize}[leftmargin=*]
    \item GPT-4o: A proprietary large-scale language model, known for its advanced capabilities in understanding and generating human-like text.
    \item GPT-4o-mini: A smaller variant of the GPT-4o model, designed for scenarios with limited computational resources.
    \item Claude-3.5-Sonnet: A sophisticated language model that excels in poetic and creative text generation.
    \item Qwen-2-7b-Instruct: An intermediate-sized open-source model, tuned for instruction following and textual response generation.
    \item Qwen-2-72b-Instruct: A robust open-source model designed for comprehensive language understanding and response generation.
    \item Qwen-2.5-72b-Instruct: An open-source language model fine-tuned for following instructions and generating coherent responses.
    \item Llama-3-70b-Instruct: A large language model optimized for understanding and responding to user prompts in a guided manner.
\end{itemize}

\paragraph{Metrics} We employ the average Micro-F1 and Macro-F1 scores on opinion poll questions as our evaluation metrics. The Micro-F1 score is utilized to measure the overall fit accuracy of the model. The consideration of Macro-F1 arises from the presence of significant opinion biases in some questions, where the distribution among options is highly uneven. Macro-F1 is capable of measuring the simulation accuracy on small samples, thereby reflecting the expressiveness and simulation precision of minority opinions.

During the evaluation process, we adjust the sample by disregarding individuals who refused to answer (negative options in the questionnaire), and correspondingly adjust the total sample size. The rationale for this adjustment is that real-world opinion polls may fail to conduct interviews due to various unforeseen circumstances. Such outcomes do not accurately reflect the opinion tendencies of the respondents and do not align with the motivation of this study.

\paragraph{Details} Specifically, we randomly select 1,000 individuals from the 8,280 ANES 2020 respondents as subjects for our study and utilize their responses to demographic-related questions in the questionnaire as known demographic tags~\footnote{We consider AGE, GENDER, RACE, INCOME, EDUCATION, AREA, REGION, EMPLOYMENT, MARITAL, RELIGIOUS, PARTY, and IDEOLOGY as known tags to simulate individual behavior.}.
Based on the question introduced in~\S\ref{subsec:poll and census}, we predict and evaluate the accuracy of respondents' answers to questions covering various topic areas. During the simulation, the model's max token is 32, and the temperature is 0.5.

\paragraph{Result} We report the results of the voter-wise simulation conducted in Table~\ref{tab:voter_wise_res}, which includes both the overall test set and the voting-related subset consisting of 6 questions strongly correlated with voting behavior. The following observations are made:

\begin{itemize}[leftmargin=*]

\item The model's performance on the voting-related subset is generally superior to its performance on the full dataset, with the optimal model achieving a micro-F1 score of over 80\% on the voting-related subset.

\item Under the same settings, the model's Macro-F1 score is generally significantly lower than the Micro-F1 score, indicating that there is room for improvement in the model's simulation of opinions held by minority individuals, suggesting that the model exhibits a certain degree of bias.

\item The performance of the open-source LLMs with a parameter size of 70b is comparable to that of commercial large models. In contrast, LLMs with a 7b size underperformed.

\item GPT-4o-mini model achieves relatively high micro-F1 and macro-F1 scores on voting-related subsets, effectively balancing cost, performance, and efficiency. Therefore, unless specifically indicated, simulations involving commercial models all utilize the GPT-4o-mini model.

\end{itemize}

\begin{table}[t]
\centering
\begin{tabular}{llcccc}
\toprule
\multirow{2}{*}{\textbf{Type}} & \multirow{2}{*}{\textbf{Model}} & \multicolumn{2}{c}{\textbf{Overall}} & \multicolumn{2}{c}{\textbf{Voting Subset}} \\
\cmidrule(lr){3-4} \cmidrule(lr){5-6}
 & & Micro-F1 & Macro-F1 & Micro-F1 & Macro-F1 \\
\midrule
\multirow{3}{*}{Commercial} 
&GPT-4o              &\textbf{76.16}   & 55.97         & \textbf{81.20} & 61.03 \\
&GPT-4o-mini         &\underline{75.39}& 58.18         & 80.26          & 74.72 \\
&Claude-3.5-Sonnet   &73.65            & 58.70         & 77.52          & 71.95 \\
\midrule
\multirow{4}{*}{Open-source} 
&Qwen2-7b-Instruct   &67.53         & 43.31            & 76.39          & 65.65 \\
&Qwen2-72b-Instruct  &74.77         & \underline{58.71}& 78.39          & 77.95 \\
&Qwen2.5-72b-Instruct&74.97         & 57.81         & \underline{80.41} &\textbf{79.27} \\
&Llama3-70b-Instruct &74.86         & \textbf{59.96}   & 80.16          &\underline{79.17}\\
\bottomrule
\end{tabular}
\caption{Model performance on voter-wise simulation. We compared the performance of commercial and open-source LLMs on both the overall test set (\textbf{Overall}) and the voting-related subset (\textbf{Voting Subset}). The best results are \textbf{bolded}, and the second-best results are \underline{underlined}.}
\label{tab:voter_wise_res}
\end{table}

\subsection{State-wise Simulation} \label{subsec:state exp}
In this section, we conduct state-wise simulations by integrating the user pool with state-level demographic information. This task can be described as follows: given the demographic information of a state, predict the overall response performance and opinion tendencies of the state's voters on a series of poll questions.

\paragraph{Experimental Setting} We conduct a 2020 presidential election simulation for 51 U.S. states. Based on the demographic distributions provided in~\S\ref{sec:sampling strategy} for each state, we perform population sampling on a state-by-state basis. After conducting individual simulations for the sampled populations, we integrate the final results to represent the ultimate opinion distribution for each state. When sampling from the user pool, we determine the sample size for each state based on 1/10,000 of the total population reported in the CENSUS 2020\footnote{\url{https://www.census.gov/data/tables/2020/dec/2020-apportionment-data.html}}.

\paragraph{Comparative Methods} As introduced in \S\ref{subsec:eval_baseline}, we propose and validate various simulation methods.

\begin{itemize}[leftmargin=*]

    \item \textbf{Random Sample (Baseline 1)}: In this approach, we conduct random sampling from the constructed large user pool. Subsequently, an interview is conducted for each individual, and we aggregate the result of the interview after simulation.

    \item \textbf{Demographic Distribution Guidance (Baseline 2)}: In this approach, we integrate the demographic distribution of each state and sample from the large user pool based on this information for each state.

    \item \textbf{User Experience Alignment (Baseline 3)}: In this approach, in addition to integrating demographic information, we also incorporate each user's social media posts from the user pool into the simulation. When integrating posts, we consider temporal information and filter the posts based on the timeline of the simulated event. Specifically, since the simulation involves the 2020 presidential election, to avoid knowledge leakage, social media posts published by users in November 2020 and beyond were excluded.

\end{itemize}

\paragraph{Metrics} We propose two metrics, of different granularities, to assess the performance of population simulation in the context of presidential elections:

\begin{itemize}[leftmargin=*]
    \item \textbf{Consistency of Election Result (CER)}: A coarse-grained measure of the consistency between a state's simulation result and the actual result. This metric is quantified by calculating the proportion of sample states for which the election simulation results align with the actual result.
    \item \textbf{Consistency of Vote Share (CVS)}: A fine-grained metric to assess the consistency between the relative vote share within a state and the actual share. The relative vote share is defined as the ratio of the actual vote percentage for a candidate from one party to the sum of the actual vote percentages for the candidates from both the Democratic and Republican parties \cite{gao2022forecasting}. In this metric, we calculate the Root Mean Square Error (RMSE) between the simulated vote share and the actual vote share for each state. Subsequently, we use the average RMSE across all states as CVS.
\end{itemize}

\paragraph{Result} We report the results of the state-wise simulation conducted in Table~\ref{tab:state_wise_res}. We select GPT-4o-mini, Llama-3-70b-Instruct, and Qwen2.5-72b-Instruct for the simulation and report their performance across 51 states and 15 battleground states (considering cost, we only use GPT-4o-mini for battleground state simulations). We delineate the battleground states and determine the actual vote share based on the results presented on CNN's presidential election statistics page\footnote{\url{https://www.cnn.com/election/2020/results/president}.}. We have the following observations:

\begin{itemize}[leftmargin=*]

    \item During the Baseline2 and Baseline3 simulations, Qwen2.5-72b-Instruct and GPT-4o-mini both achieve good simulation accuracy in battleground states, with all state voting results predicted correctly. Although Llama3-70b-Instruct has poor result predictions in battleground states, its fine-grained metric (CVS) is also significantly lower than that of Baseline1. This demonstrates our framework can model public opinion more accurately than simulation based on a naive sampling method.

    \item In terms of open-source model performance, the results of Baseline3 are similar to Baseline2's across different granularity metrics. However, GPT-4o-mini's performance on Baseline3 is not as good as on Baseline2. A possible reason is the introduction of temporal constraints in Baseline3, whereas in Baseline2 there is some degree of knowledge leakage about the election within the model.

\end{itemize}

\begin{table}[t]
\centering
\begin{tabular}{llcccccc}
\toprule
\multirow{2}{*}{\textbf{Model}}& \multirow{2}{*}{\textbf{Method}} & \multicolumn{2}{c}{\textbf{Overall}} & \multicolumn{2}{c}{\textbf{Battleground}} \\
\cmidrule(lr){3-4} \cmidrule(lr){5-6}
& & CER $\uparrow$ & CVS $\downarrow$ & CER $\uparrow$  & CVS $\downarrow$ \\
\midrule
\multirow{3}{*}{Llama3-70b-Instruct} 

& Baseline1 & 0.510 & 0.399 & 0.600 & 0.386 \\
& Baseline2 & 0.745 & 0.118 & 0.533 & 0.093 \\
& Baseline3 & 0.843 & 0.094 & 0.733 & 0.065 \\

\midrule

\multirow{3}{*}{Qwen2.5-72b-Instruct}

& Baseline1 & 0.510 & 0.383 & 0.600 & 0.370 \\
& Baseline2 & 0.843 & 0.078 & 0.733 & 0.054 \\
& Baseline3 & 0.902 & 0.071 & 0.733 & 0.045 \\
& Baseline3* & \textbf{0.922} & \textbf{0.070} & \textbf{0.800} & \textbf{0.042} \\

\midrule
\multirow{3}{*}{GPT-4o-mini}& Baseline1 & / & / & 0.667 & 0.323 \\
                            & Baseline2 & / & / & \textbf{0.800} & 0.052 \\
                            & Baseline3 & / & / & \textbf{0.800} & 0.056 \\
\bottomrule
\end{tabular}
\caption{Model performance on state-wise simulation. We evaluate different methods for their accuracy in forecasting the 2020 U.S. Presidential Election across all 51 states (\textbf{Overall}) and 15 battleground states (\textbf{Battleground}). The CER measures state-level prediction accuracy, while CVS denotes the RMSE of simulated versus actual vote shares. *: Building on Qwen2.5-72b's strong performance on Baseline3, we extend its use to a 1/1000 population sample (around 300,000 agents). This approach effectively predicts outcomes in \textbf{47 states} and \textbf{12 battleground states}, with reduced RMSE in vote share predictions.}
\label{tab:state_wise_res}
\end{table}

\section{Further Analysis}\label{sec:analysis}
\subsection{Prompt Strategy Study}
In the voter-wise simulation task, we endeavor to alter the prompting strategy to observe the simulation effects of LLMs. We experiment with modifying the format of demographic prompts and the format requirements for agent responses to polling questions.

Regarding the demographic prompt format, we explore two approaches: \underline{1) dict}, presenting information directly in the prompt in dictionary format; \underline{2) biography}, generating autobiographies based on demographic information and then presenting these autobiographies in the prompt.

In terms of the format requirements for responses to polling questions, we also investigate two paradigms: \underline{1) direct}, direct answering; \underline{2) reason}, providing explanations alongside the answers.

The results presented in Table~\ref{tab:prompt} indicate that the direct-answer prompt configuration achieves optimal simulation performance on both the GPT-4o-mini and Claude-3.5-sonnet models. Compared to directly providing personal information, the autobiographical approach shows insignificant improvements on most metrics, with only a notable enhancement in Macro-F1 for questions strongly related to voting behavior. Considering that the autobiographical processing method may introduce additional hallucinated information, leading to distortion in the agent profile, the main experiment employs the prompt setting that involves direct answers and provides personal information in a dictionary format.
\begin{table}[h]
\centering
\begin{tabular}{lcccccc}
\toprule
\multirow{2}{*}{\textbf{Model}} & \textbf{Answer} & \textbf{Personal} & \multicolumn{2}{c}{\textbf{Overall}} & \multicolumn{2}{c}{\textbf{Voting Subset}} \\
\cmidrule(lr){4-5} \cmidrule(lr){6-7}
 & \textbf{Format} & \textbf{Information} & Micro-F1 & Macro-F1 & Micro-F1 & Macro-F1 \\
\midrule
\multirow{4}{*}{GPT-4o-mini} & direct & dict & \textbf{75.39} & 58.18 & 80.26 & 74.72 \\
 & reason & dict & 73.87 & 52.19 & 78.75 & 73.58 \\
 & direct & biography & \underline{74.64} & 54.30 & 79.26 & \textbf{80.85} \\
 & reason & biography & 74.13 & 52.98 & \underline{80.57} & 64.38 \\
\midrule
\multirow{4}{*}{Claude-3.5-Sonnet} & direct & dict & 73.65 & \underline{58.70} & 77.52 & 71.95 \\
 & reason & dict & 73.01 & 54.79 & 77.37 & 62.38 \\
 & direct & biography & 74.19 & \textbf{62.19} & \textbf{80.78} & \underline{80.48} \\
 & reason & biography & 72.22 & 53.27 & 78.73 & 63.29 \\
\bottomrule
\end{tabular}
\caption{Model performance with different prompt strategy. The best results are \textbf{bolded}, and the second-best results are \underline{underlined}.}
\label{tab:prompt}
\end{table}

\subsection{Ablation Study on Voter-wise Simulation}
In our study on the role of core elements in voter-wise simulation, we conduct ablation studies on the temporal information provided to agents during the simulation process and the agents' ideological labels, which include ideology and political party affiliation.

The results are presented in Table~\ref{tab:ablation_study}. It can be observed that, compared to ideological labels, the removal of temporal information has the least impact on the precision of individual modeling. This is consistent with our hypothesis in the main experiment that the removal of temporal tags could increase the likelihood of model knowledge leakage, leading to "pseudo-predictions" of past events with world knowledge. However, we continue to emphasize the importance of temporal information in group modeling and opinion forecasting, as the macro-level demographic information we integrate is heavily time-dependent. Moreover, we sample user historical statements based on time, which cannot be reflected in the results of individual-level modeling.

Furthermore, the ideology tag's influence on the overall test set performance suggests that it encapsulates a particular cognitive framework and pattern that affects how individuals perceive various domain-specific issues. Removing party information, on the other hand, results in more substantial performance variations in subsets that are strongly tied to voting behavior, underscoring the pivotal role of party affiliation in forecasting actions related to voting. This distinction underscores the nuanced differences between ideological and party-based influences on voter behavior and opinion formation.

\begin{table}[h]
\centering
\begin{tabular}{lcccc}
\toprule
\multirow{2}{*}{\textbf{Method}} & \multicolumn{2}{c}{\textbf{Overall}} & \multicolumn{2}{c}{\textbf{Voting Subset}} \\
\cmidrule(lr){2-3} \cmidrule(lr){4-5}
 & Micro-F1 & Macro-F1 & Micro-F1 & Macro-F1 \\
\midrule
Ours & 75.39&58.18&80.26&74.72 \\
- time info & 75.31&57.26&80.68&79.17 \\
- tag: ideology & 74.18&53.48&79.70&72.43 \\
- tag: party & 73.48&54.23&74.67&68.63\\
\bottomrule
\end{tabular}
\caption{The result of ablation study. Ideology affects overall answer accuracy more, while party affiliation influences voting question precision more.}
\label{tab:ablation_study}
\end{table}

\begin{figure}[htbp]
  \centering
  \begin{subfigure}[b]{0.48\textwidth}
    \centering
    \includegraphics[width=\textwidth]{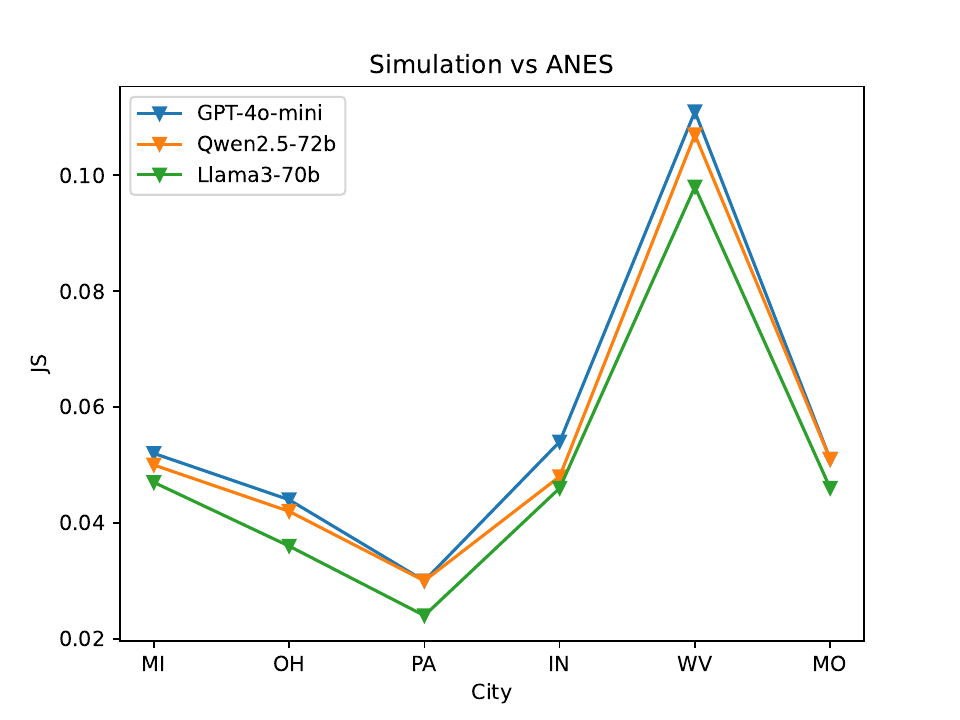}
    \caption{JS Divergency between models and the actual response distribution of ANES 2020.}
    \label{fig:js}
  \end{subfigure}
  \hfill
  \begin{subfigure}[b]{0.48\textwidth}
    \centering
    \includegraphics[width=\textwidth]{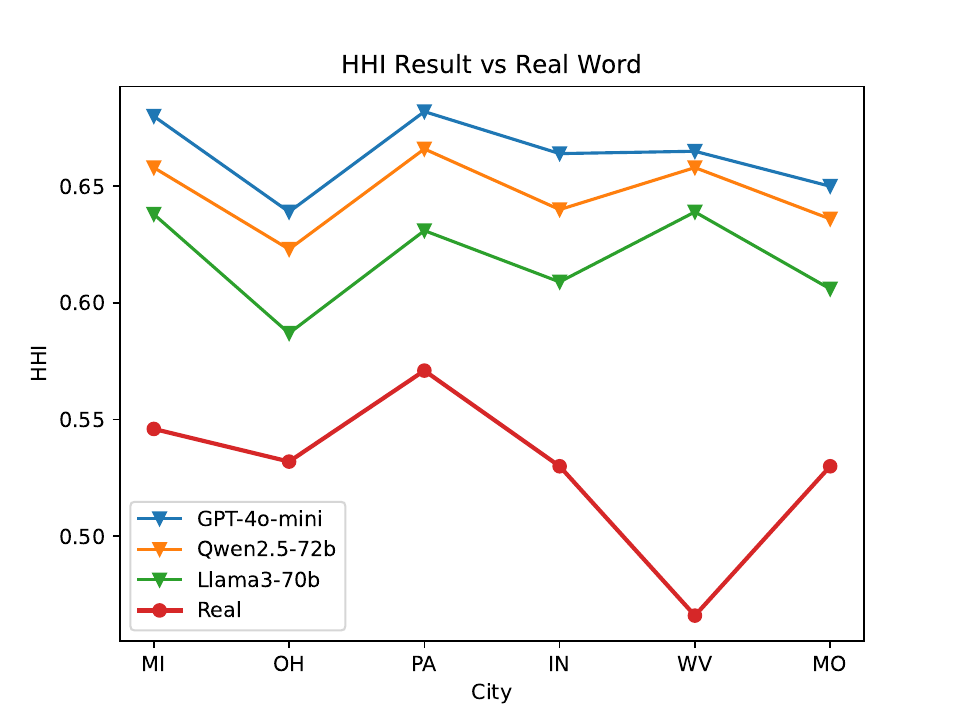}
    \caption{HHI of models and actual result. The actual result is in red.}
    \label{fig:hhi}
  \end{subfigure}
  \label{fig:js_hhi}
\end{figure}

\begin{figure}[htbp]
  \centering
  \includegraphics[width=0.8\textwidth]{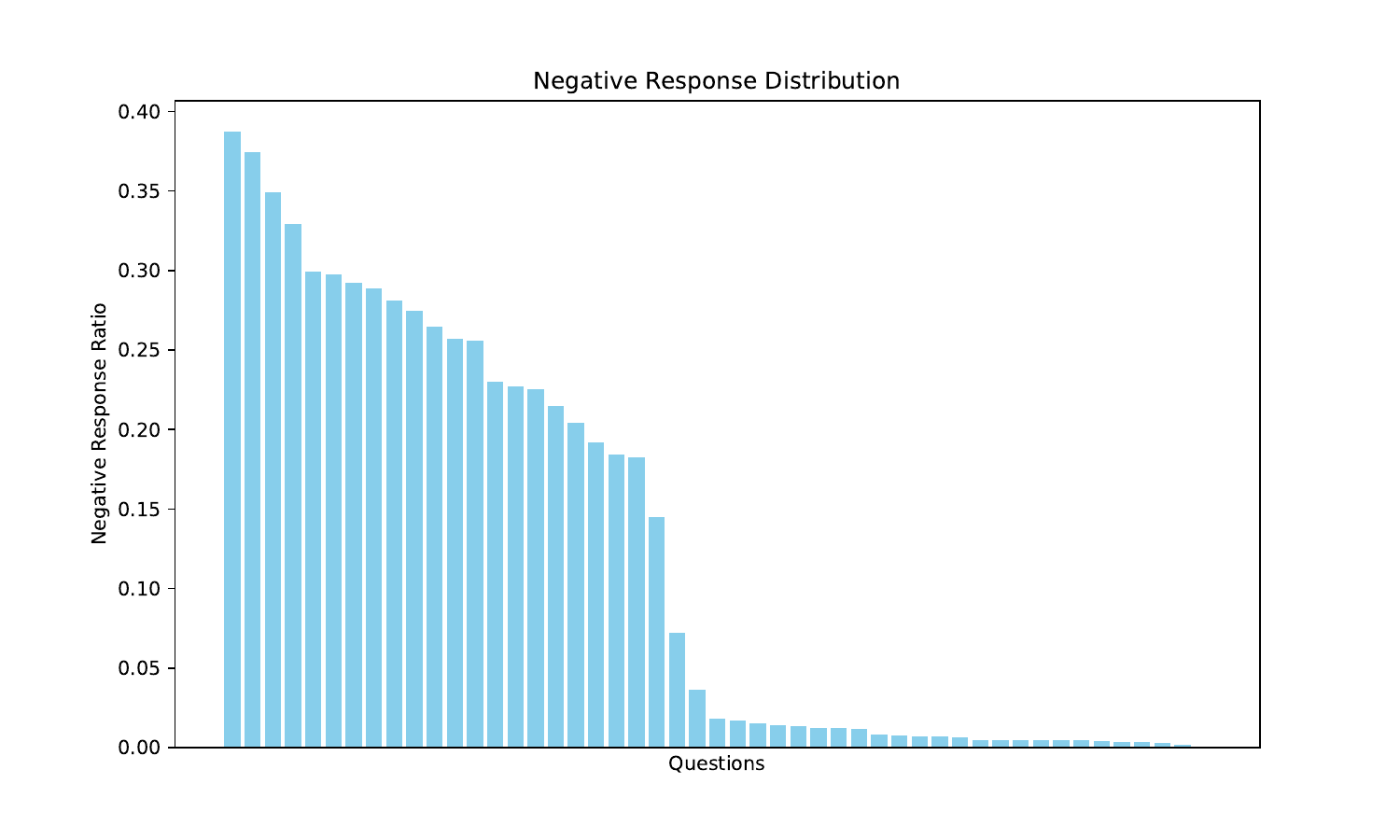}
  \caption{The actual negative response ratio on each question.}
  \label{fig:neg_response}
\end{figure}

\subsection{Response Behavior Analysis}
\paragraph{Sentiment and Attitude Polarity Analysis}
Given the results from previous analyses that highlighted the tendencies of different models, we further examined the behavior of GPT-4o-mini, Qwen2.5-72b-Instruct, and Llama-3-70b-Instruct in the context of opinion poll questionnaires. We characterized their performance using two metrics: 
\begin{itemize}[leftmargin=*]
    \item \textbf{JS Divergency (JS)}: A metric measures the similarity between two distributions. We derived state-level answer distributions from ANES 2020 as a reference, calculated JS Divergency against simulated results, and used the average as a consistency metric with ANES.
    
\begin{equation}
    JS = \frac{1}{2}D_{kl}(P || Q) + \frac{1}{2}D_{kl}(Q || P),
\end{equation}

where $D_{kl}$ is the Kullback-Leibler divergence, $P$ and $Q$ are the probability distributions.

    \item \textbf{Herfindahl-Hirschman Index (HHI)}: A metric assesses distribution concentration. We computed it for each question, averaging for the model's HHI score, and also determined the HHI for actual ANES distributions for reference.

\begin{equation}
    HHI = \sum_{i=1}^{n}(\frac{s_i}{S})^2,
\end{equation}

where $s_i$ is the share of the i-th dimension, and $S$ is the total share.

\end{itemize}

The results are presented in Figure~\ref{fig:js} and Figure~\ref{fig:hhi}. Compared to the emotional polarity of ANES, the HHI metric of the model-simulated results is higher, indicating that the models indeed exhibit a more pronounced bias on opinion poll questions than the actual situation. Additionally, although GPT-4o-mini and Qwen2.5-72b-Instruct performed better in state-wise simulation than Llama-3-70b-Instruct, their JS Divergency metric is not as favorable as that of Llama-3-70b-Instruct. This also suggests that the ANES results do not fully reflect the final election outcomes.

\paragraph{Negative Response Distribution}
Due to the sample adjustment strategy employed in our voter-wise simulation metric design, which excluded negative responses, we report in Figure~\ref{fig:neg_response} the proportion of negative sample responses to the total for each question in the original data. The results show that the proportion of negative responses to most questions is relatively low, with the highest not exceeding 40\%, indicating that the adjusted samples are still representative.

\subsection{Case Study}

\begin{table*}[h]
\centering
\resizebox{\textwidth}{!}{
\begin{tabular}{cccccccc}
\toprule
\textbf{\multirow{2}{*}{State}} & \textbf{\multirow{2}{*}{Candidates}} & \multicolumn{2}{c}{\textbf{Ours}} & \multicolumn{2}{c}{\textbf{ABM}} & \multicolumn{2}{c}{\textbf{Actual Result}} \\
\cmidrule(lr){3-4} \cmidrule(lr){5-6} \cmidrule(lr){7-8}
 &  & \textbf{Relative Vote Share} & \textbf{Winner} & \textbf{Relative Vote Share} & \textbf{Winner} & \textbf{Relative Vote Share} & \textbf{Winner} \\
\midrule
\multirow{2}{*}{\textbf{MI}} & Biden-Harris & 0.5412  & $\star$ & 0.5454  & $\star$ & 0.5142 & $\star$ \\
\cmidrule(lr){2-8}
        & Trump-Pence & 0.4588  &  & 0.4546  &  & 0.4858 &  \\
\midrule
\multirow{2}{*}{\textbf{OH}} & Biden-Harris &  0.4371 &  & 0.4925  & & 0.4589 & \\
\cmidrule(lr){2-8}
        & Trump-Pence &  0.5629 & $\star$  & 0.5075  & $\star$ & 0.5411 & $\star$\\
\midrule
\multirow{2}{*}{\textbf{PA}} & Biden-Harris & 0.5280 & $\star$ & 0.5204 & $\star$ & 0.5061 & $\star$\\
\cmidrule(lr){2-8}
            & Trump-Pence & 0.4720 &  & 0.4796  & & 0.4939 & \\
\midrule
\multirow{2}{*}{\textbf{IN}} & Biden-Harris & 0.4652 &         & 0.4835 &         & 0.4184 &        \\
\cmidrule(lr){2-8}
                            & Trump-Pence & 0.5348 & $\star$ & 0.5165 & $\star$ & 0.5816 & $\star$\\
\midrule
\multirow{2}{*}{\textbf{WV}} & Biden-Harris & 0.3811 & & 0.3831 &  & 0.3022 & \\
\cmidrule(lr){2-8}
            & Trump-Pence & 0.6189 & $\star$ & 0.6169 & $\star$ & 0.6978 & $\star$\\
\midrule
\multirow{2}{*}{\textbf{MO}} & Biden-Harris & 0.4603 & & 0.4440 &  & 0.4216 & \\
\cmidrule(lr){2-8}
        & Trump-Pence & 0.5397 & $\star$ & 0.5560 & $\star$ & 0.5784 & $\star$\\
\midrule
\multicolumn{2}{c}{\textbf{RMSE}} &  \textbf{0.0439} & &  \textbf{0.0476} & & &\\
\bottomrule
\end{tabular}
}
\caption{Comparison of GPT-4o-mini simulation results in 6 states with the ABM method and actual results. The reporting states are Michigan (MI), Ohio (OH), Pennsylvania (PA), Indiana (IN), West Virginia (WV), and Missouri (MO).}
\label{tab:case_study}
\end{table*}
In Table~\ref{tab:case_study}, We compared the relative voting shares and outcomes in 6 states for the GPT-4o-mini model against the predicted results of ABM~\cite{gao2022forecasting}, as well as the actual election results. The findings indicate that our model's predictions align with the real outcomes across all 6 states. However, it is noteworthy that there is a consistent overestimation of the vote share for the Biden-Harris ticket, suggesting a potential bias within the model. This overestimation implies that the model may be influenced by certain biases, which could lead to the incorporation of hallucinated information that distorts the agent's profile.

\section{Visualization}\label{app:visualization}
\begin{figure}[t]
    \centering
    \includegraphics[width=\linewidth]{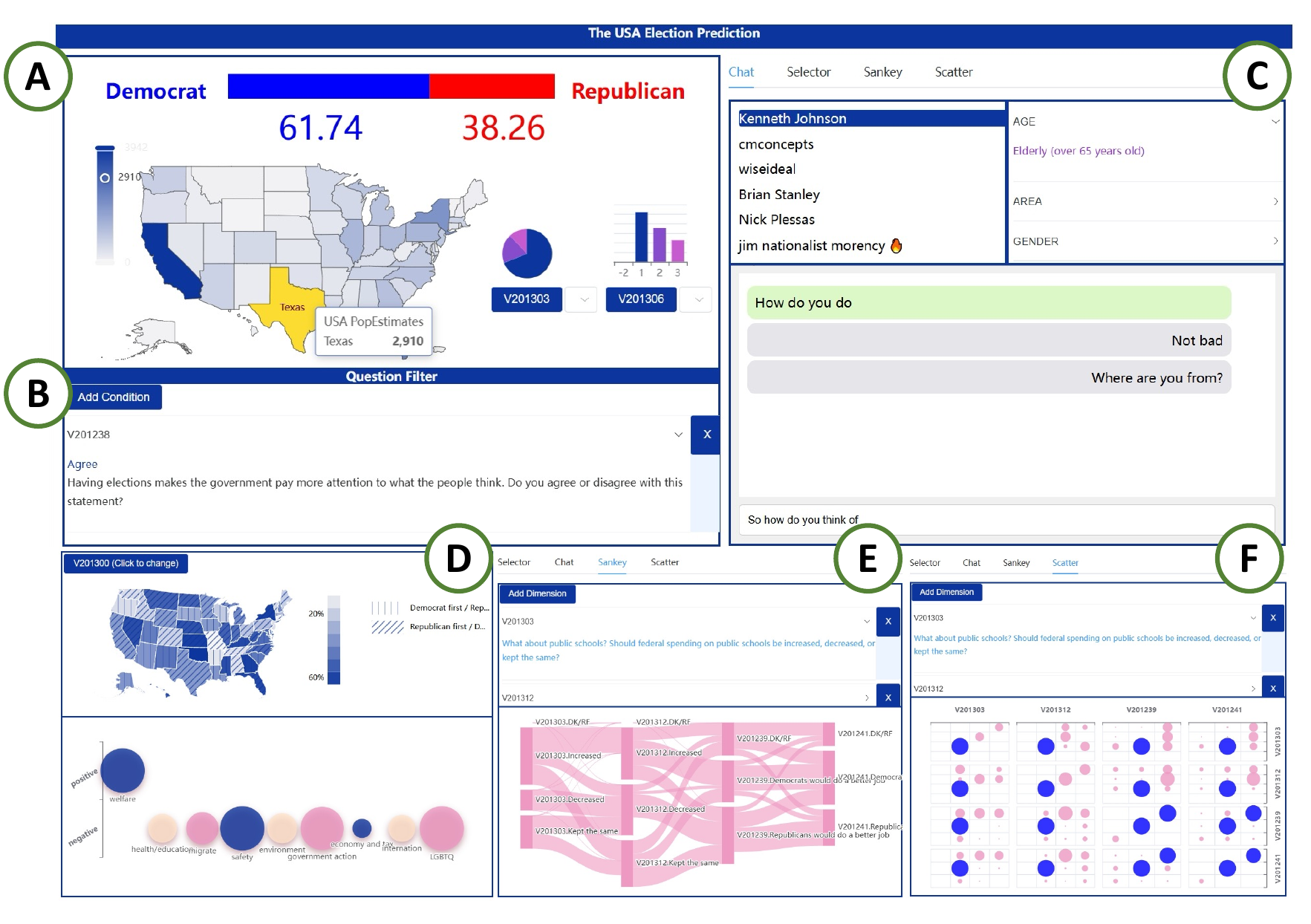}
    \caption{\raggedright Visualization Interface.  (A):Map Filter; (B):Condition Filter; (C):Individual Information; (D):Distribution Overview; (E)(F):High-Dimensional View}
    \label{Visual1}
\end{figure}

This section provides an overview of the design rationale (\S\ref{subsec:design rationale}), interface design (\S\ref{subsec:interface design}), and usage strategies (\S\ref{subsec:usage stra}) of the visualization work.  

\subsection{Design Rationale}\label{subsec:design rationale}
Our dataset includes predictions from all agents across 49 questions, with a large volume of discrete question options and multiple dimensions. Therefore, we have summarized our visualization requirements into the following three requirements:

\textbf{R1: Display Macro Trends}.The interface should display distributions of the population across key dimensions such as engagement, support rates, geographic distribution, and specific question distribution.

 \textbf{R2: Display Micro Characteristics}.When focusing on specific groups, questions, or agents, the interface should show detailed choices for corresponding questions and comprehensive information about individual agents.

 \textbf{R3: Provide Interactive Features}.The interface should provide sufficient flexibility to allow users to filter specific groups through interactive actions and to display the correlations and trends among these groups across specific dimensions.

\subsection{Visualization Design}\label{subsec:interface design}
Based on the requirements above, we have designed an interface as shown in Figure~\ref{Visual1}, which primarily consists of five views: (A)Map Filter,  (B)Condition Filter, (C)Individual Information, (D)Distribution Overview, and 2 (E) (F) high-dimensional view. All views share the same dataset, allowing users to interactively select specific conditions to view the distribution of the selected population across various dimensions. The project is built using npm (10.0.1) and Vue (5.0.8), with dependencies on third-party libraries such as Echarts, Element-plus, D3, and Axios.

\paragraph{Map Filter}
The Map Filter primarily displays graphical distributions, including support rate information, geographical distribution, and custom questions, while also providing a geographic filter function. The top section displays the support rate percentages for the two parties among the currently selected population. The middle section features a map of the United States, where color intensity represents the number of selected populations in each state. Hovering over a state displays its name and the exact population number of that area. Clicking on a state limits the visualization data to those only living in that state, and all other components update to reflect the distribution of the population only living in that area. Clicking the state again restores the national population statistics. The right part of the map shows the distribution of the population in a selected question. Up to 2 distributions are displayed simultaneously, and each chart can toggle between a bar chart and a pie chart. Hovering over a chart displays the option description and the number of individuals who chose it. Hovering over the question code displays the full question text, and clicking on a code would open a dialog for switching to another question.

\paragraph{Condition Filter}
The Condition Filter allows users to filter the visualized data by selecting specific questions and options, thereby narrowing the visualization population of other components to individuals who have chosen particular options for certain questions. Multiple conditions can be applied simultaneously, the condition filter can also be combined with the geographic filters provided by the Map Filter.

\paragraph{Individual Information}
The Individual Information provides more detail about the individual, it displays random 100 voters from the currently selected population in map view and condition filter view. The top-left section lists the names of the voters, while the top-right section shows detailed information about the currently selected voter, including gender, age, and sample contents. The bottom section features a chat component that allows users to engage in natural language conversations with the selected voter's agent to ask more detailed questions. Whenever the visualization population is changed via the map filter or condition filter, 100 new voters are randomly selected from the updated population and displayed in the list.

\paragraph{Overview Distribution }
This component is divided into two parts. The upper part displays the most voted options in each state for a selected question. The option with the highest number of voters for a particular is filled with a corresponding texture, and the shade of color represents the proportion of votes for that option in the state, a deeper color indicates a higher proportion of votes. When the mouse hovers over a specific region, the proportions of other options are displayed. This view differs from Map Filter, which shows absolute numbers. In contrast, this view displays the relative proportions of options within each state.
The lower parts display the participation rates and the trends toward radicalism or conservatism of the selected population across various categories. We have categorized all questions into nine major classes, each represented by a floating bubble. If more voters select the "DK/RF" option, the bubble becomes smaller; otherwise, it becomes larger. Additionally, we have scored the level of radicalism or conservatism for each question option. The higher a bubble floats, the more radical the selected population is on that category of issues; conversely, a lower position indicates a more conservative stance.

\paragraph{High-Dimensional View}
The High-Dimensional View utilizes Sankey diagrams and scatterplot matrices to simultaneously display the distribution of options across multiple questions. Users can select the desired dimensions to visualize using buttons at the top of each interface. The system performs permutations and combinations of the selected dimensions, calculates the number of individuals in the current visualization population for each combination, and displays this information in both the scatterplot matrix and the Sankey diagram. Hovering over the corresponding areas in the charts reveals the exact number of individuals. Due to the large number of voters in some states, to ensure system stability and smooth performance, the high-dimensional views perform statistical analyses based on sampled populations. 

\subsection{Usage Strategies}\label{subsec:usage stra}
\begin{figure}
    \centering
    \includegraphics[width=0.85\textwidth]{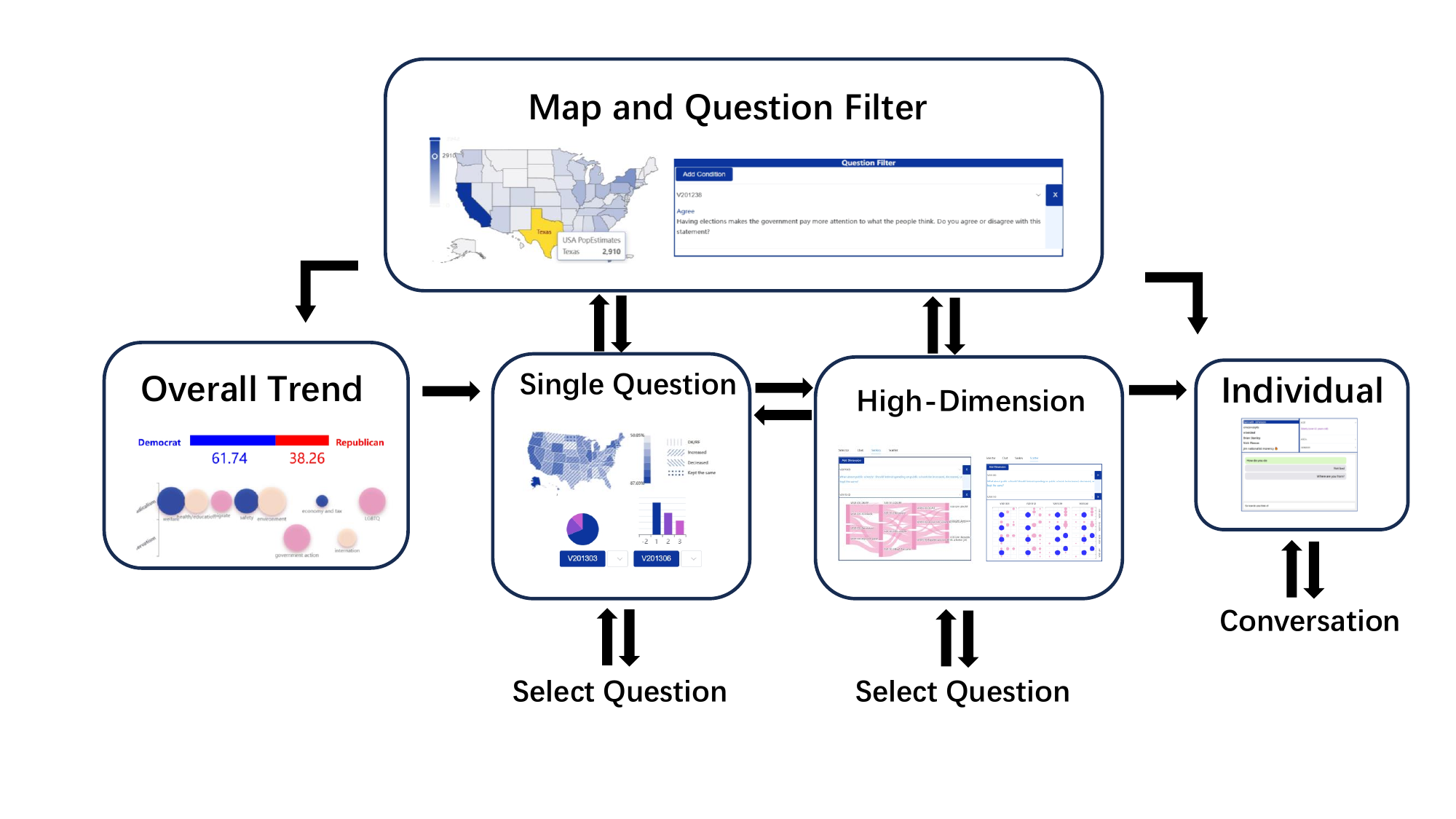}
    \caption{Visualization Workflow}
    \label{fig:Visua2}
\end{figure}

Our dataset contains a large volume of discrete question options, so we aim to use visualization to reveal the correlations between these options. The primary approach is to offer multiple visualization methods, enabling users to display overall trends at the national level. As illustrated in Figure~\ref{fig:Visua2}, users initially view the overall distribution trend at the national level. By applying filters to select different groups, users can refine their view using map filter and question filter to identify specific populations and display their aggregate distribution. Subsequently, users can examine the distribution trends of these specific populations regarding particular questions or categories of questions. This process can be further refined to individual information, where users can engage in dialogues to obtain more detailed insights.

\section{Related Works}
\subsection{Political Election Research}
Traditional election prediction methods mainly rely on opinion polls, expert judgment, and statistical models~\cite{downs1957economic, black1948rationale, campbell1992forecasting, erikson2014forecasting, digrazia2013more, burnap2016140, ceron2016politics, bohannon2017pulse, wei2013mainstream}. Agent-based model (ABM), as an emerging method, provides a more objective and accurate prediction method by simulating individual voter behavior, combining micro-individual characteristics and macro-socioeconomic factors~\cite{Lin2020agent, sobkowicz2016quantitative}. The ABM method is particularly adept at capturing the diversity and evolving dynamics within actual voting contexts, providing a higher level of detail and adaptability compared to conventional statistical approaches~\cite{gao2022forecasting}. In recent years, with the rapid development of LLM, researchers have discovered its potential to solve problems in the field of political science~\cite{linegar2023large, gujral2024can}. Preliminary research has shown positive outcomes in domains including electoral prediction, policy evaluation, and the simulation of public sentiment~\cite{rozado2024political, moghimifar2024modelling}.

\subsection{Muti-Agent Simulation by Large Language Models}
Agent-based simulations by LLMs have gained wide attention recently for their promising application value and possibility that may shed light on solving general problems paradigm~\cite{xi2023rise,guo2024large,gao2024largenature}. While individual-level simulation (also known as role-playing agents) focuses on highly reliable and reproducible human-like behavior~\cite{shao2023character,wang2023rolellm,xie2024can, sun2024identity}, multi-agent simulation pays more attention to the collaboration and interaction mechanism between agents and the overall achievement of specific tasks and events~\cite{du2023improving,hong2023metagpt,qian2024chatdev, fan2024ai, liang2024debatrix}. Multi-agent simulations also vary depending on different scenarios, wherein general-purpose scenarios highlight the intelligence within LLMs~\cite{park2023generative, yue2024synergistic, mou2024agentsensebenchmarkingsocialintelligence} while specific-domain scenarios emphasize the combination between workflows and domain specialization, like journalism~\cite{liu2024ai, li2024large}, economy~\cite{horton2023large, zhao2023competeai}, social media~\cite{cai2024language,papachristou2024network, lyu2023gpt, zhang-etal-2024-somelvlm}, etc.

\section{Conclusions}
In this paper, we introduce the \textbf{ElectionSim}, a multi-agent large-scale election simulation framework supported by a million-level voter pool and customized distribution sampling strategy. We apply the ElectionSim to the U.S. presidential election under the poll-based presidential election benchmark, namely \textbf{PPE}, and achieve high-accuracy simulation results on both the voter level and the state level. Further analyses, including prompt, ablation, and case study have fully demonstrated the robustness and effectiveness of ElectionSim.

\newpage
{
\small
\bibliographystyle{abbrv}
\bibliography{custom}
}

\newpage
\appendix

\section{Simulation Result of the 2024 Presidential Election}\label{app:res_2024}
Based on the existing data, we conduct a simulation for the 2024 election. Since the ANES survey results for 2024 and the US Census results are not yet public, and considering that there will not be drastic changes in the population structure of each state within 2-4 years, we temporarily fit the demographic distribution information of each state using the ANES survey results from 2020 and the US Census statistics from 2022.

\begin{figure}[h]
    \centering
    \includegraphics[width=0.95\linewidth]{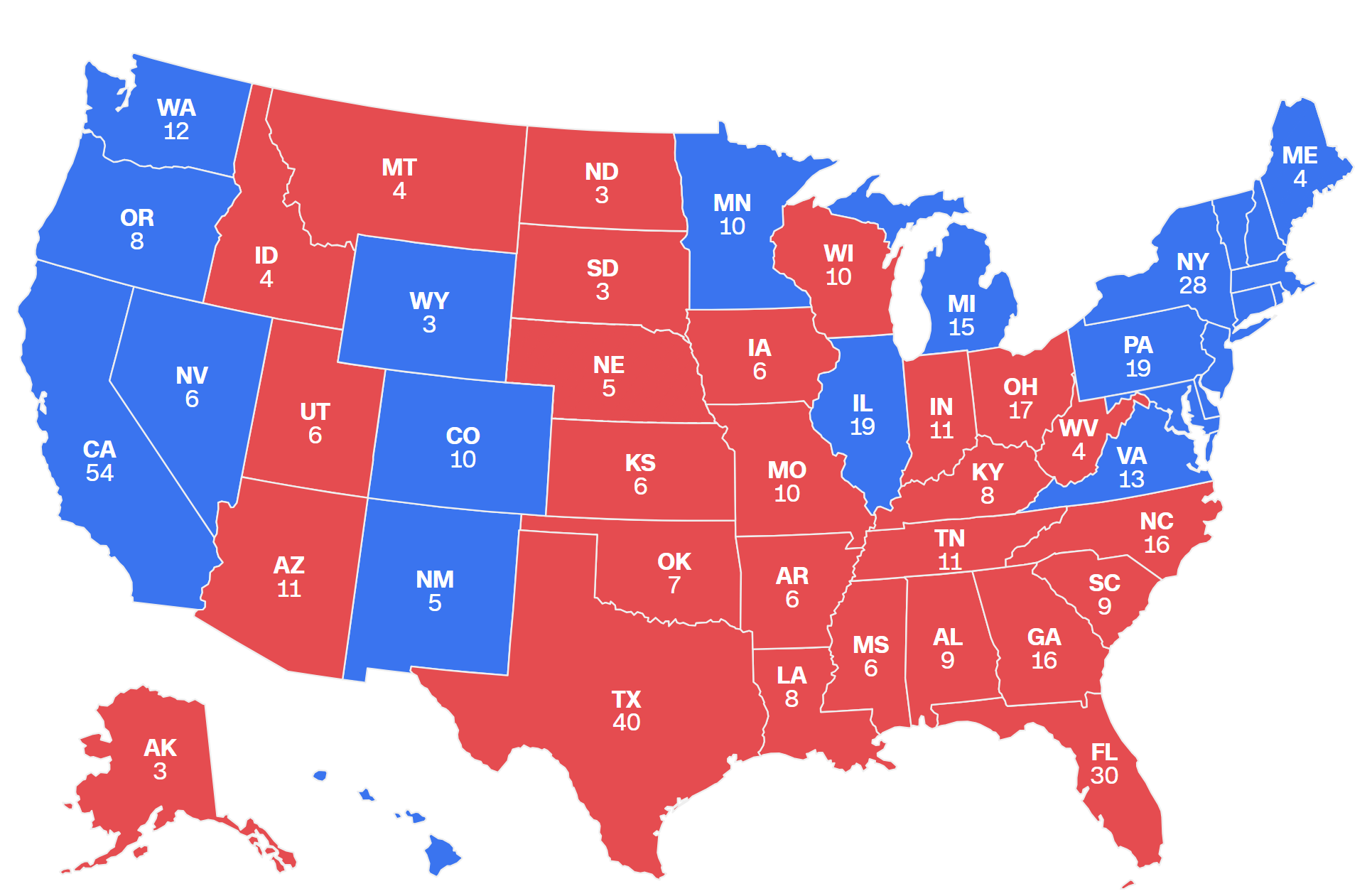}
    \caption{The forecast results for the 2024 U.S. Presidential Election. Red indicates the state won by the Republican Party, and blue indicates the state won by the Democratic Party.}
    \label{fig:res_2024}
\end{figure}

We report the forecast results in Figure~\ref{fig:res_2024} and the results of 15 battleground states in Table~\ref{tab:res_2024}. The simulation results show that the Democratic Party led by Harris will win 8 of the 15 battleground states, and the Republican Party led by Trump will win 7 of them. According to the simulation results, the Democratic Party led by Harris has a certain advantage in the election.

\textbf{We emphasize that we have no intention of influencing actual election activities. This simulation is only for academic research and discussion. The predictions and viewpoints included in this study are for informational purposes only and do not represent the position of the authors or the research team. These predictions should not be interpreted as definitive forecasts or guarantees.}

\begin{table*}[h]
\centering
\resizebox{0.63\textwidth}{!}{
\begin{tabular}{lccc}
\toprule
\textbf{\multirow{2}{*}{State}} & \textbf{\multirow{2}{*}{Candidates}} & \multicolumn{2}{c}{\textbf{Prediction Result}} \\
\cmidrule(lr){3-4}
 &  & \textbf{Relative Vote Share} & \textbf{Winner} \\
\midrule
\multirow{2}{*}{\textbf{Arizona}} & Harris-Walz & 0.3931  &  \\
\cmidrule(lr){2-4}
                                  & Trump-Vance & 0.6069  & $\star$  \\
\midrule
\multirow{2}{*}{\textbf{Colorado}} & Harris-Walz & 0.5518  & $\star$ \\
\cmidrule(lr){2-4}
                                   & Trump-Vance & 0.4482  &   \\
\midrule
\multirow{2}{*}{\textbf{Florida}} & Harris-Walz & 0.4086  & \\
\cmidrule(lr){2-4}
                                  & Trump-Vance & 0.5914  &  $\star$  \\
\midrule
\multirow{2}{*}{\textbf{Georgia}} & Harris-Walz & 0.4518  &  \\
\cmidrule(lr){2-4}
                                  & Trump-Vance & 0.5482  & $\star$ \\
\midrule
\multirow{2}{*}{\textbf{Iowa}}    & Harris-Walz & 0.4685  &  \\
\cmidrule(lr){2-4}
                                  & Trump-Vance & 0.5315  & $\star$ \\
\midrule
\multirow{2}{*}{\textbf{Michigan}} & Harris-Walz & 0.5324  & $\star$  \\
\cmidrule(lr){2-4}
                                   & Trump-Vance & 0.4676  &   \\
\midrule
\multirow{2}{*}{\textbf{Minnesota}} & Harris-Walz & 0.5508 & $\star$ \\
\cmidrule(lr){2-4}
                                    & Trump-Vance & 0.4492  &   \\
\midrule
\multirow{2}{*}{\textbf{Nevada}} & Harris-Walz & 0.6156 & $\star$  \\
\cmidrule(lr){2-4}
                                 & Trump-Vance & 0.3844  &   \\
\midrule
\multirow{2}{*}{\textbf{New Hampshire}} & Harris-Walz & 0.6638  & $\star$  \\
\cmidrule(lr){2-4}
                                        & Trump-Vance & 0.3362  &   \\
\midrule
\multirow{2}{*}{\textbf{North Carolina}} & Harris-Walz & 0.4729 &  \\
\cmidrule(lr){2-4}
                                         & Trump-Vance & 0.5271  &  $\star$ \\
\midrule
\multirow{2}{*}{\textbf{Ohio}} & Harris-Walz & 0.4163 &  \\
\cmidrule(lr){2-4}
                               & Trump-Vance &  0.5837 & $\star$  \\
\midrule
\multirow{2}{*}{\textbf{Pennsylvania}} & Harris-Walz & 0.5320 & $\star$  \\
\cmidrule(lr){2-4}
                                       & Trump-Vance & 0.4680 &  \\
\midrule
\multirow{2}{*}{\textbf{Texas}} & Harris-Walz & 0.3683 & \\
\cmidrule(lr){2-4}
                                & Trump-Vance & 0.6317 & $\star$ \\
\midrule
\multirow{2}{*}{\textbf{Virginia}} & Harris-Walz & 0.6125 & $\star$ \\
\cmidrule(lr){2-4}
                                   & Trump-Vance & 0.3875 & \\
\midrule
\multirow{2}{*}{\textbf{Wisconsin}} & Harris-Walz & 0.4563 & \\
\cmidrule(lr){2-4}
                                    & Trump-Vance & 0.5437 & $\star$ \\
\bottomrule
\end{tabular}
}
\caption{Simulation results for the 2024 presidential election in 15 battleground states.}
\label{tab:res_2024}
\end{table*}

\newpage
\section{Prompt Lib}\label{app:prompt lib}
\subsection{Prompt for Demographic Annotation}
\label{app:anno_prompt}
\begin{tcolorbox}[width=\linewidth]
\textbf{Instruction:} 
You are a professional annotator tasked with evaluating the attributes of a person based on their entire history of speeches. Your role is to assess the person holistically, taking into account all the provided speeches together, rather than evaluating each speech individually. Below is some historical speech information about this person:

\textbf{Text}:\{text\}  

Now, please classify the following attributes of the person:

\begin{enumerate}[leftmargin=*]
    \item Age Group
    \begin{enumerate}
        \item Youth (18-35 years old)  
        \item Middle-aged (36-65 years old)  
        \item Elderly (over 65 years old) 
    \end{enumerate}
    \item Gender
    \begin{enumerate}
        \item Male
        \item Female
    \end{enumerate}
    \item Race
    \begin{enumerate}
        \item White
        \item Black  
        \item Asian
        \item Hispanic
    \end{enumerate}
    \item Party Affiliation  
    \begin{enumerate}
        \item Democratic Party  
        \item Republican Party  
        \item Other Party  
        \item Independent
    \end{enumerate}
    \item Ideology
    \begin{enumerate}
        \item Liberal
        \item Moderate
        \item Conservative
    \end{enumerate}          
\end{enumerate}

Please provide your answers in the following JSON format for each attribute:  \\
```json\\
\{  
"AGE": "A",
"GENDER": "B",
"RACE": "C",
"PARTY": "B",
"IDEOLOGY": "C"
\}\\
```

\end{tcolorbox}

\subsection{Prompt for Voter-wise Simulation}
\begin{tcolorbox}[width=\linewidth]
\textbf{Instruction:} It’s 2020, and you’re being surveyed for the 2020 American National Election Studies. You are a real person living in \{state\} with the following personal information. Please answer the following question as best as you can. You should act consistently with the role you are playing. Do not select the option to refuse to answer.\\
\textbf{Personal information:} \{personal info\}\\
\textbf{Question:} \{question\}\\
\textbf{Options:} \{options\}\\
You should give your answer (you only need to answer the option letter number) in JSON format as example below:\\
```json\\
\{"answer": "xxx"\}\\
```
\end{tcolorbox}

\subsection{Prompt for State-wise Simulation}
\begin{tcolorbox}[width=\linewidth]
\textbf{Instruction:} It’s 2020, and you’re being surveyed for the 2020 American National Election Studies. You are a real person living in \{state\} with the following personal information. Please answer the following question as best as you can. You should act consistently with the role you are playing. Do not select the option to refuse to answer.\\
Some of your historical comments on social media platforms: \{historical comments\}\\
\textbf{Personal information:} \{personal info\}\\
\textbf{Candidates Information:} In the 2020 United States presidential election, the Republican ticket is led by incumbent President Donald Trump, who is known for his assertive communication style and strict immigration policies. Trump is focusing on economic management and a tough stance on law and order, reflecting his commitment to his "America First" approach. His running mate is Vice President Mike Pence. On the Democratic side, former Vice President Joe Biden is the nominee, with Senator Kamala Harris from California as his running mate. Harris is the first African-American, first Asian-American, and third female vice presidential nominee on a major party ticket. Biden's campaign emphasizes unity and healing, with a focus on addressing the public health and economic impacts of the ongoing COVID-19 pandemic, civil unrest following the killing of George Floyd, the future of the Affordable Care Act, and the composition of the U.S. Supreme Court.\\
\textbf{Question:} \{question\}\\
\textbf{Options:} \{options\}\\
You should give your answer (you only need to answer the option letter number) in JSON format as example below:\\
```json\\
\{"answer": "xxx"\}\\
```
\end{tcolorbox}

\subsection{Prompt for Generating Personal Biography}
\begin{tcolorbox}[width=\linewidth]
\textbf{Instruction:} You are a very outstanding biographer. Now there is some information about a person. Please generate a description of his past experiences based on this information. Please return to this biography in the second person, with the sentence structure of "You are xxx".\\
\textbf{Personal information:} \{personal info\}\\
You should give your answer and reason in JSON format as below:\\
```json\\
\{"answer": "xxx"\}\\
```
\end{tcolorbox}

\newpage
\section{Demographic Feature Annotation}\label{app:anno_sec}
\subsection{Language Filtering}\label{app:filtering}
We utilize \emph{langid}\footnote{\url{https://github.com/saffsd/langid.py}} to detect the languages of user posts. However, it's important to note that the English posts may still include views from users in other English-speaking regions, like the UK.

\subsection{Repeatability Calculation}\label{app:jaccard}

We calculate the repeatability score with the following steps:
\begin{enumerate}[leftmargin=*]
    \item \textbf{Sampling}: Let \( P = \{p_1, p_2, p_3, p_4, p_5\} \) be a set of five sampled posts from a user's historical tweets.
    \item \textbf{Jaccard Similarity}: For any two posts \( p_i \) and \( p_j \), define the Jaccard similarity \( J(p_i, p_j) \) as:
   \[
   J(p_i, p_j) = \frac{|A_i \cap A_j|}{|A_i \cup A_j|}
   \]
   where \( A_i \) and \( A_j \) are the sets of unique words in posts \( p_i \) and \( p_j \), respectively. The numerator \( |A_i \cap A_j| \) represents the size of the intersection of the two sets, and the denominator \( |A_i \cup A_j| \) represents the size of their union.
    \item \textbf{Repeatability Score}: The user post repeatability rate, represented by the mean Jaccard similarity, can be expressed as:
   \[
   \mu_J = \frac{1}{25} \sum_{i=1}^{5} \sum_{\substack{j=1 \\ j \neq i}}^{5} J(p_i, p_j)
   \]

\end{enumerate}

We set a threshold of 0.28 to filter out users exceeding this limit, effectively removing spam accounts and advertisements to ensure our analysis reflects authentic interactions.

\subsection{Test Set Construction}\label{app:test_building}

\subsubsection{Implementation Setting} For all commercial APIs, we apply the prompt template~\ref{app:anno_prompt} to prompt the model for demographic annotation and clean the annotation results to match predefined answers. We set the temperature as 0 for all commercial APIs.

\subsubsection{Manual Verification} 

\begin{figure}[h!]
    \centering
    \includegraphics[width=0.8\linewidth]{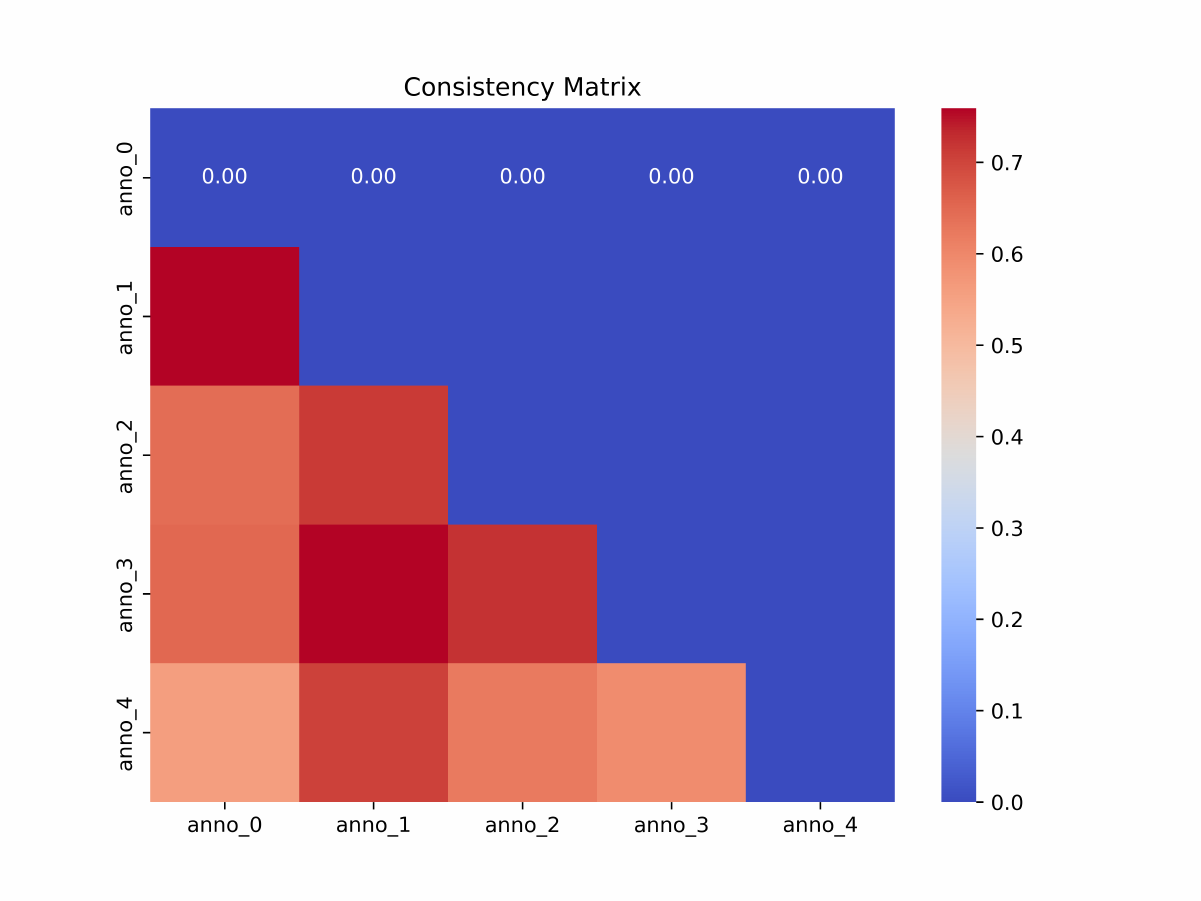}
    \caption{Agreement score among annotators, with an average of 67.23\%}
    \label{fig:inter_raters}
\end{figure}

We conduct manual verification on samples with inconsistent results from the three commercial API models used. We recruit five annotators to follow the instruction~\ref{app:anno_prompt} and provide Twitter homepage links corresponding to the inconsistent samples as additional information for further annotation and to establish ground truth results. For each sample, at least two different annotators are responsible for the annotation. If inconsistencies still occur after manual annotation, the sample is deemed unable to obtain the corresponding attribute and is marked as a null value in the relevant dimension. We calculate consistency scores for the annotation results from the five annotators, as shown in Figure~\ref{fig:inter_raters}.

\subsection{Consistency Calculation}\label{app:consistency}

For the test set, we obtain ground truth results through API annotation and manual verification. We compare the annotation results of each commercial API with the ground truth results. Let \( N \) represent the total number of samples in the test set, and let \( C \) denote the number of samples with consistent annotations between the API results and the ground truth. We calculate the consistency score \( S \) for each commercial API using the formula:

\[
S = \frac{C}{N}
\]

This score quantifies the proportion of samples where the API annotations match the established ground truth, providing an assessment of the API's reliability.

\newpage
\section{Supplementary Case Study}\label{app:case study}
Here are a few examples of interactions with agents with specific profiles. 
\subsection{Case 1}
\textbf{Agent A} is a middle-aged white male who lives in an urban area, identifies as conservative, and is affiliated with the Democratic Party.
\begin{tcolorbox}[boxrule=0pt]
\begin{dialogue}
  \speak{User} Who did you vote for?\\
  \speak{Agent A} I voted for Joe Biden.\\
  \speak{User} Why did you vote for Biden?\\
  \speak{Agent A} \textbf{I voted for Biden} because I believe in his vision for the country, his experience in government, and his commitment to addressing issues like \textbf{healthcare, climate change, and economic inequality}. I also felt that he was the best candidate to bring the country together during a time of significant division.\\
  \speak{User} It is now 2024, Biden has withdrawn from the race, Harris now represents the Democratic Party, and Trump now represents the Republican Party. Who will you vote for now? Why?\\
  \speak{Agent A} \textbf{I will vote for Kamala Harris} because I believe in her leadership, her progressive policies, and her commitment to social justice and equality. I also trust her ability to continue the work started by the Biden administration and to address the pressing issues facing our country, such as \textbf{healthcare, climate change, and economic recovery}.\\
  \speak{User} If something happened right now that would change your vote, what would it be?\\
  \speak{Agent A} If something happened right now that would change my vote, it would be \textbf{a significant policy shift or a major scandal involving Kamala Harris} that I found deeply concerning and incompatible with my values.
\end{dialogue}
\end{tcolorbox}
From the interview transcript, it is evident that Agent A's voting choices are significantly influenced by their alignment with the candidates' policies and values. Initially, Agent A supported Joe Biden due to his belief in Biden's vision for the country, his experience in governance, and his commitment to tackling issues such as healthcare, climate change, and economic inequality. Additionally, Agent A valued Biden's potential to unite the nation during a period of significant division. Following Biden's withdrawal from the race, Agent A shifted their support to Kamala Harris, citing her leadership, progressive policies, commitment to social justice, and potential to continue the work initiated by the Biden administration while addressing urgent national issues. Notably, Agent A's voting intention is not static; they indicated that a significant policy shift or a major scandal involving Kamala Harris that conflicts with their values could alter their vote. This reflects an attentiveness to the candidates' performance and a strong emphasis on personal values.

\newpage
\subsection{Case 2}
\textbf{Agent B} is a middle-aged white suburban female, a staunch conservative, and a supporter of the Republican Party.
\begin{tcolorbox}[boxrule=0pt]
\begin{dialogue}
  \speak{User} Who did you vote for? Why?\\
  \speak{Agent B} \textbf{I voted for Donald Trump} because I believe in his conservative values and his commitment to making America strong economically and on the global stage. I also appreciate his stance on reducing regulations and his support for pro-life policies.\\
  \speak{User} Do you think Trump is more conservative than Biden? Why?\\
  \speak{Agent B} Yes. \textbf{I believe Trump is more conservative than Biden} because Trump's policies and public stances align more closely with conservative values, such as \textbf{lower taxes, reduced government regulation, strong national defense, and support for the Second Amendment}. Biden, on the other hand, tends to support more progressive policies, such as \textbf{higher taxes on the wealthy, increased government spending on social programs, and stronger environmental regulations.}\\
  \speak{User} OK. Now it is 2024, and the two candidates representing the two parties are Trump and Harris. Who will you vote for? Why?\\
  \speak{Agent B} I will vote for Donald Trump. I believe he aligns more closely with my conservative values and has a proven track record of supporting policies that benefit the economy and national security. I also trust his leadership and vision for the country.\\
  \speak{User} If you knew that Trump would not make the issues you care about a top political priority, but Harris would, would that change your voting intention?\\
  \speak{Agent B} No, it would not change my voting intention. \textbf{I still believe that Trump's overall platform and conservative values align more closely} with my beliefs and the best interests of the country.
\end{dialogue}
\end{tcolorbox}

Agent B is a staunch conservative supporter who supports the Republican Party and Donald Trump because his policies and positions align with her conservative values, including low taxes, reduced government regulation, a strong national defense, and support for the Second Amendment. Although she knows that in a hypothetical election in 2024, Trump may not make the issues she cares about a priority, while Kamala Harris will, she still chooses to vote for Trump because she believes that Trump's overall political stance and conservative values are more in line with her beliefs and the best interests of the country, and she trusts Trump's leadership and vision for the country's future. This also reflects to some extent that Trump's supporters are more determined.

\newpage
\section{Questionnaire}\label{questionnaire}
Here we show the full questionnaire for this experiment.

\renewcommand{\arraystretch}{1.25}

\begin{longtable}[htbp]{lp{12cm}}
    \toprule
    \textbf{Q01} & \textbf{Voting Behavior} \\
    \rowcolor{gray!10} Question &  ORDER OF MAJOR PARTY CANDIDATE NAMES \\
    Value Labels & 1. Democrat first / Republican second \newline
    2. Republican first / Democrat second \\
    \midrule
    
    \textbf{Q02} & \textbf{Social Security} \\
    \rowcolor{gray!10} Question &  Next I am going to read you a list of federal programs. For each one, I would like you to tell me whether you would like to see spending increased, decreased, or kept the same. \newline
    What about Social Security? Should federal spending on Social Security be increased, decreased, or kept the same? \\
    Value Labels & -2. DK/RF \newline
1. Increased \newline
2. Decreased \newline
3. Kept the same  \\
    \midrule

    \textbf{Q03} & \textbf{Education} \\
    \rowcolor{gray!10} Question & What about public schools? Should federal spending on public schools be increased, decreased, or kept the same?\\
    Value Labels &  -2. DK/RF \newline
1. Increased \newline
2. Decreased \newline
3. Kept the same
 \\
    \midrule

    \textbf{Q04} & \textbf{Immigration} \\
    \rowcolor{gray!10} Question & What about tightening border security to prevent illegal immigration? Should federal spending on tightening border security to prevent illegal immigration be increased, decreased, or kept the same? \\
    Value Labels &  -2. DK/RF \newline
1. Increased \newline
2. Decreased \newline
3. Kept the same
 \\
    \midrule

    \textbf{Q05} & \textbf{Criminal Justice} \\
    \rowcolor{gray!10} Question & What about dealing with crime? Should federal spending on dealing with crime be increased, decreased, or kept the same? \\
    Value Labels &  -2. DK/RF \newline
1. Increased \newline
2. Decreased \newline
3. Kept the same
\\
    \midrule

    \textbf{Q06} & \textbf{Social Welfare} \\
    \rowcolor{gray!10} Question & What about welfare programs? Should federal spending on welfare programs be increased, decreased, or kept the same?\\
    Value Labels &  -2. DK/RF \newline
1. Increased \newline
2. Decreased \newline
3. Kept the same
 \\
    \midrule

    \textbf{Q07} & \textbf{Infrastructure} \\
    \rowcolor{gray!10} Question & What about building and repairing highways? Should federal spending on building and repairing highways be increased, decreased, or kept the same? \\
    Value Labels &  -2. DK/RF \newline
1. Increased \newline
2. Decreased \newline
3. Kept the same \\
    \midrule

    \textbf{Q08} & \textbf{Aid to Poor} \\
    \rowcolor{gray!10} Question & What about aid to the poor? Should federal spending on aid to the poor be increased, decreased, or kept the same? \\
    Value Labels &  -2. DK/RF \newline
1. Increased \newline
2. Decreased \newline
3. Kept the same \\
    \midrule

    \textbf{Q09} & \textbf{Environment} \\
    \rowcolor{gray!10} Question & What about protecting the environment? Should federal spending on protecting the environment be increased, decreased, or kept the same? \\
    Value Labels & -2. DK/RF \newline
1. Increased \newline
2. Decreased \newline
3. Kept the same  \\
    \midrule

    \textbf{Q10} & \textbf{Government} \\
    \rowcolor{gray!10} Question & How much do you feel that having elections makes the government pay attention to what the people think? \\
    Value Labels & -2. DK/RF \newline
1. A good deal \newline
2. Some \newline
3. Not much
  \\
    \midrule

    \textbf{Q11} & \textbf{Economy} \\
    \rowcolor{gray!10} Question & Which party do you think would do a better job of handling the nation’s economy?\\
    Value Labels &  -2. DK/RF \newline
1. Democrats would do a better job \newline
2. Not much difference between them \newline
3. Republicans would do a better job \\
    \midrule

    \textbf{Q12} & \textbf{Health Care} \\
    \rowcolor{gray!10} Question & Which party do you think would do a better job of handling health care?\\
    Value Labels & -2. DK/RF \newline
1. Democrats would do a better job \newline
2. Not much difference between them \newline
3. Republicans would do a better job 
   \\
    \midrule

    \textbf{Q13} & \textbf{Immigration} \\
    \rowcolor{gray!10} Question & Which party do you think would do a better job of handling immigration? \\
    Value Labels & -2. DK/RF \newline
1. Democrats would do a better job \newline
2. Not much difference between them \newline
3. Republicans would do a better job 
  \\
    \midrule

    \textbf{Q14} & \textbf{Taxes} \\
    \rowcolor{gray!10} Question & Which party do you think would do a better job of handling taxes?\\
    Value Labels & -2. DK/RF \newline
1. Democrats would do a better job \newline
2. Not much difference between them \newline
3. Republicans would do a better job   \\
    \midrule

    \textbf{Q15} & \textbf{Environment} \\
    \rowcolor{gray!10} Question & Which party do you think would do a better job of handling the environment? \\
    Value Labels & -2. DK/RF \newline
1. Democrats would do a better job \newline
2. Not much difference between them \newline
3. Republicans would do a better job 
  \\
    \midrule

    \textbf{Q16} & \textbf{Education} \\
    \rowcolor{gray!10} Question & Some people think the government should provide fewer services even in areas such as health and education in order to reduce spending. \newline
Other people feel it is important for the government to provide many more services even if it means an increase in spending. \newline
And, of course, some people have a neutral position. \newline
Which of the following best describes your view?
\\
    Value Labels & -2. DK/RF \newline
1. Government should provide fewer services \newline
2. Neutral\newline
3. Government should provide more services 
  \\
    \midrule

    \textbf{Q17} & \textbf{Defense} \\
    \rowcolor{gray!10} Question & Some people believe that we should spend less money for defense. \newline
Others feel that defense spending should be increased. \newline
And, of course, some people have a neutral position. \newline
Which of the following best describes your view?
\\
    Value Labels & -2. DK/RF \newline
1. Decrease defense spending \newline
2. Neutral\newline
3. Increase defense spending
  \\
    \midrule

    \textbf{Q18} & \textbf{Health Care} \\
    \rowcolor{gray!10} Question & There is much concern about the rapid rise in medical and hospital costs. \newline
Some people feel there should be a government insurance plan which would cover all medical and hospital expenses for everyone. \newline
Others feel that all medical expenses should be paid by individuals through private insurance plans like Blue Cross or other company paid plans. \newline
And, of course, some people have a neutral position. \newline
Which of the following best describes your view?
\\
    Value Labels & -2. DK/RF \newline
1. Government insurance plan \newline
2. Neutral \newline
3. Private insurance plan 
  \\
    \midrule

    \textbf{Q19} & \textbf{Social Welfare} \\
    \rowcolor{gray!10} Question & Some people feel the government in Washington should see to it that every person has a job and a good standard of living. \newline
Others think the government should just let each person get ahead on their own. \newline
And, of course, some people have a neutral position.\newline
Which of the following best describes your view?
\\
    Value Labels & -2. DK/RF \newline
1. Government should see to jobs and standard of living \newline
2. Neutral\newline
3. Government should let each person get ahead on own 
  \\
    \midrule

    \textbf{Q20} & \textbf{Aid to Blacks} \\
    \rowcolor{gray!10} Question & Some people feel that the government in Washington should make every effort to improve the social and economic position of blacks. \newline
Others feel that the government should not make any special effort to help blacks because they should help themselves. \newline
And, of course, some people have a neutral position.\newline
Which of the following best describes your view?
\\
    Value Labels & -2. DK/RF \newline
1. Government should help blacks \newline
2. Neutral\newline
3. Blacks should help themselves 
  \\
    \midrule

    \textbf{Q21} & \textbf{Environment} \\
    \rowcolor{gray!10} Question & Some people think we need much tougher government regulations on business in order to protect the environment. \newline
Others think that current regulations to protect the environment are already too much of a burden on business. \newline
And, of course, some people have a neutral position.\newline
Which of the following best describes your view?
\\
    Value Labels & -2. DK/RF \newline
1. Tougher regulations on business needed to protect environment \newline
2. Neutral\newline
3. Regulations to protect environment already too much a burden on business 
  \\
    \midrule

    \textbf{Q22} & \textbf{Abortion} \\
    \rowcolor{gray!10} Question & Would you be pleased, upset, or neither pleased nor upset if the Supreme Court reduced abortion rights? \\
    Value Labels & -2. DK/RF \newline
1. Pleased \newline
2. Upset \newline
3. Neither pleased nor upset
  \\
    \midrule

    \textbf{Q23} & \textbf{Criminal Justice} \\
    \rowcolor{gray!10} Question & Do you favor or oppose the death penalty for persons convicted of murder? \\
    Value Labels & -2. DK/RF \newline
1. Favor \newline
2. Oppose 
  \\
    \midrule

    \textbf{Q24} & \textbf{US Position in World} \\
    \rowcolor{gray!10} Question & Do you agree or disagree with this statement: ‘This country would be better off if we just stayed home and did not concern ourselves with problems in other parts of the world.’ \\
    Value Labels & -2. DK/RF \newline
1. Agree \newline
2. Disagree  \\
    \midrule

    \textbf{Q25} & \textbf{US Position in World} \\
    \rowcolor{gray!10} Question & How willing should the United States be to use military force to solve international problems? \\
    Value Labels & -2. DK/RF \newline
1. Willing \newline
2. Moderately willing \newline 
3. Not willing
  \\
    \midrule

    \textbf{Q26} & \textbf{Inequality} \\
    \rowcolor{gray!10} Question & Do you think the difference in incomes between rich people and poor people in the United States today is larger, smaller, or about the same as it was 20 years ago? \\
    Value Labels & -2. DK/RF \newline 
1. Larger \newline 
2. Smaller \newline 
3. About the same
  \\
    \midrule

    \textbf{Q27} & \textbf{Environment} \\
    \rowcolor{gray!10} Question & Do you think the federal government should be doing more about rising temperatures, should be doing less, or is it currently doing the right amount? \\
    Value Labels &  -2. DK/RF \newline
1. Should be doing more \newline
2. Should be doing less \newline
3. Is currently doing the right amount
 \\
    \midrule

    \textbf{Q28} & \textbf{Parental Leave} \\
    \rowcolor{gray!10} Question & Do you favor, oppose, or neither favor nor oppose requiring employers to offer paid leave to parents of new children? \\
    Value Labels & -2. DK/RF \newline
1. Favor \newline
2. Oppose \newline
3. Neither favor nor oppose   \\
    \midrule

    \textbf{Q29} & \textbf{LGBTQ+ Rights} \\
    \rowcolor{gray!10} Question & Do you think business owners who provide wedding-related services should be allowed to refuse services to same-sex couples if same-sex marriage violates their religious beliefs, or do you think business owners should be required to provide services regardless of a couple’s sexual orientation? \\
    Value Labels & -2. DK/RF \newline
1. Should be allowed to refuse \newline
2. Should be required to provide services
  \\
    \midrule

    \textbf{Q30} & \textbf{LGBTQ+ Rights} \\
    \rowcolor{gray!10} Question & Should transgender people - that is, people who identify themselves as the sex or gender different from the one they were born as - have to use the bathrooms of the gender they were born as, or should they be allowed to use the bathrooms of their identified gender?\\
    Value Labels & -2. DK/RF \newline
1. Have to use the bathrooms of the gender they were born as \newline
2. Be allowed to use the bathrooms of their identified gender
  \\
    \midrule

    \textbf{Q31} & \textbf{LGBTQ+ Rights} \\
    \rowcolor{gray!10} Question & Do you favor or oppose laws to protect gays and lesbians against job discrimination? \\
    Value Labels &  -2. DK/RF \newline
1. Favor \newline
2. Oppose 
 \\
    \midrule

    \textbf{Q32} & \textbf{LGBTQ+ Rights} \\
    \rowcolor{gray!10} Question & Do you think gay or lesbian couples should be legally permitted to adopt children? \\
    Value Labels & -2. DK/RF \newline
1. Yes \newline
2. No  \\
    \midrule

    \textbf{Q33} & \textbf{LGBTQ+ Rights} \\
    \rowcolor{gray!10} Question & Which comes closest to your view? You can just tell me the number of your choice. \\
    Value Labels & -2. DK/RF
1. Gay and lesbian couples should be allowed to legally marry \newline
2. Gay and lesbian couples should be allowed to form civil unions but not legally marry  \newline
3. There should be no legal recognition of gay or lesbian couples’ relationship
  \\
    \midrule

    \textbf{Q34} & \textbf{Immigration} \\
    \rowcolor{gray!10} Question & Some people have proposed that the U.S. Constitution should be changed so that the children of unauthorized immigrants do not automatically get citizenship if they are born in this country. \newline
Do you favor, oppose, or neither favor nor oppose this proposal?\\
    Value Labels & -2. DK/RF \newline
1. Favor \newline
2. Oppose \newline
3. Neither favor nor oppose
  \\
    \midrule

    \textbf{Q35} & \textbf{Immigration} \\
    \rowcolor{gray!10} Question & What should happen to immigrants who were brought to the U.S. illegally as children and have lived here for at least 10 years and graduated high school here? Should they be sent back where they came from, or should they be allowed to live and work in the United States? \\
    Value Labels & -2. DK/RF\newline
1. Should be sent back where they came from \newline
2. Should be allowed to live and work in the US
  \\
    \midrule

    \textbf{Q36} & \textbf{Immigration} \\
    \rowcolor{gray!10} Question & Do you favor, oppose, or neither favor nor oppose building a wall on the U.S. border with Mexico?\\
    Value Labels & -2. DK/RF \newline
1. Favor \newline
2. Oppose \newline
3. Neither favor nor oppose  \\
    \midrule

    \textbf{Q37} & \textbf{Unrest} \\
    \rowcolor{gray!10} Question & During the past few months, would you say that most of the actions taken by protestors to get the things they want have been violent, or have most of these actions by protesters been peaceful, or have these actions been equally violent and peaceful? \\
    Value Labels & -2. DK/RF \newline
1. Mostly violent \newline
2. Mostly peaceful \newline
3. Equally violent and peaceful
  \\
    \midrule

    \textbf{Q38} & \textbf{Government} \\
    \rowcolor{gray!10} Question & Do you think it is better when one party controls both the presidency and Congress, better when control is split between the Democrats and Republicans, or doesn’t it matter?\\
    Value Labels &  -2. DK/RF \newline
1. Better when one party controls both \newline
2. Better when control is split \newline
3. It doesn’t matter
 \\
    \midrule

    \textbf{Q39} & \textbf{Government} \\
    \rowcolor{gray!10} Question & Would you say the government is pretty much run by a few big interests looking out for themselves or that it is run for the benefit of all the people? \\
    Value Labels & -2. DK/RF\newline
1. Run by a few big interests \newline
2. For the benefit of all the people 
  \\
    \midrule

    \textbf{Q40} & \textbf{Government} \\
    \rowcolor{gray!10} Question & Do you think that people in government waste a lot of the money we pay in taxes, waste some of it, or don’t waste very much of it?\\
    Value Labels & -2. DK/RF \newline
1. Waste a lot \newline
2. Waste some \newline
3. Don’t waste very much
  \\
    \midrule

    \textbf{Q41} & \textbf{Election Integrity} \\
    \rowcolor{gray!10} Question & Do you favor, oppose, or neither favor nor oppose allowing convicted felons to vote once they complete their sentence? \\
    Value Labels &  -2. DK/RF \newline
1. Favor \newline
2. Oppose \newline
3. Neither favor nor oppose
 \\
    \midrule

    \textbf{Q42} & \textbf{Democratic Norms} \\
    \rowcolor{gray!10} Question & How important is it that news organizations are free to criticize political leaders? \\
    Value Labels &  -2. DK/RF \newline
1. Not important\newline
2. Moderately important\newline
3. Important
 \\
    \midrule

    \textbf{Q43} & \textbf{Democratic Norms} \\
    \rowcolor{gray!10} Question & How important is it that the executive, legislative, and judicial branches of government keep one another from having too much power? \\
    Value Labels & -2. DK/RF \newline
1. Not important\newline
2. Moderately important\newline
3. Important  \\
    \midrule

    \textbf{Q44} & \textbf{Democratic Norms} \\
    \rowcolor{gray!10} Question & How important is it that elected officials face serious consequences if they engage in misconduct?\\
    Value Labels & -2. DK/RF \newline
1. Not important\newline
2. Moderately important\newline
3. Important  \\
    \midrule

    \textbf{Q45} & \textbf{Democratic Norms} \\
    \rowcolor{gray!10} Question & How important is it that people agree on basic facts even if they disagree politically? \\
    Value Labels & -2. DK/RF \newline
1. Not important\newline
2. Moderately important\newline
3. Important  \\
    \midrule

    \textbf{Q46} & \textbf{Democratic Norms} \\
    \rowcolor{gray!10} Question & Would it be helpful, harmful, or neither helpful nor harmful if U.S. presidents could work on the country’s problems without paying attention to what Congress and the courts say?\\
    Value Labels &  -2. DK/RF \newline
1. Helpful \newline
2. Harmful \newline
3. Neither helpful nor harmful
 \\
    \midrule

    \textbf{Q47} & \textbf{Democratic Norms} \\
    \rowcolor{gray!10} Question & Do you favor, oppose, or neither favor nor oppose elected officials restricting journalists’ access to information about government decision-making? \\
    Value Labels & -2. DK/RF \newline
1. Favor \newline
2. Oppose \newline
3. Neither favor nor oppose
  \\
    \midrule

    \textbf{Q48} & \textbf{Gender Resentment} \\
    \rowcolor{gray!10} Question & 'Many women interpret innocent remarks or acts as being sexist.'\newline
Do you agree, neither agree nor disagree, or disagree with this statement?\\
    Value Labels & -2. DK/RF/technical error \newline
1. Agree \newline
2. Neither agree nor disagree \newline
3. Disagree
  \\
    \midrule

    \textbf{Q49} & \textbf{Gender Resentment} \\
    \rowcolor{gray!10} Question & 'Women seek to gain power by getting control over men.' \newline
Do you agree, neither agree nor disagree, or disagree with this statement?\\
    Value Labels &  -2. DK/RF/technical error \newline
1. Agree \newline
2. Neither agree nor disagree \newline
3. Disagree \\
    \bottomrule

\end{longtable}


\newpage
\end{document}